\documentclass[a4paper,fleqn]{cas-dc}

\usepackage[numbers]{natbib}
\usepackage{gensymb}
\usepackage{graphicx,caption}
\usepackage{circledsteps}
\usepackage{subcaption}
\def\tsc#1{\csdef{#1}{\textsc{\lowercase{#1}}\xspace}}
\tsc{WGM}
\tsc{QE}
\tsc{EP}
\tsc{PMS}
\tsc{BEC}
\tsc{DE}

\usepackage{tikz}
\newcommand*\circled[1]{\tikz[baseline=(char.base)]{
            \node[shape=circle,draw,inner sep=1pt, scale=0.85] (char) {#1};}}

\begin{document}
\setlength{\mathindent}{0pt}
\let\WriteBookmarks\relax
\def\floatpagepagefraction{1}
\def\textpagefraction{.001}
\shorttitle{Linear regression is competitive with Machine Learning in building MPC}
\shortauthors{F Bünning et~al.}

\title [mode = title]{Physics-informed linear regression is competitive with two Machine Learning methods in residential building MPC}                      

\author[1,2]{Felix Bünning}
\cormark[1]
\credit{Conceptualization, Methodology, Software, Validation, Investigation, Visualization, Supervision, Writing - Original Draft}

\author[1,2]{Benjamin Huber}
\credit{Methodology, Software, Validation, Investigation, Writing - Review \& Editing}

\author[1,2]{Adrian Schalbetter}
\credit{Methodology, Software, Validation, Investigation, Writing - Review \& Editing}

\author[2]{Ahmed Aboudonia}
\credit{Conceptualization, Supervision, Writing - Review \& Editing}

\author[2]{Mathias {Hudoba de Badyn}}
\credit{Supervision, Writing - Review \& Editing}

\author[1]{Philipp Heer}
\credit{Supervision, Funding acquisition, Writing - Review \& Editing, Conceptualization}

\author[2]{Roy S. Smith}
\credit{Supervision, Funding acquisition, Writing - Review \& Editing, Conceptualization}

\author[2]{John Lygeros}
\credit{Supervision, Funding acquisition, Writing - Review \& Editing, Conceptualization}

\address[1]{Empa, Urban Energy Systems Laboratory, Überlandstrasse 129, 8600 Dübendorf, Switzerland}
\address[2]{Automatic Control Laboratory, Department of Electrical Engineering and Information Technology, ETH Zürich, Switzerland}

\cortext[cor1]{Corresponding author}

\begin{abstract}
Because physics-based building models are difficult to obtain as each building is individual, there is an increasing interest in generating models suitable for building MPC directly from measurement data. 
Machine learning methods have been widely applied to this problem and validated mostly in simulation; there are, however, few studies on a direct comparison of different models or validation in real buildings to be found in the literature. Methods that are indeed validated in application often lead to computationally complex non-convex optimization problems.
Here we compare physics-informed Autoregressive–Moving-Average with Exogenous Inputs (ARMAX) models to Machine Learning models based on Random Forests and Input Convex Neural Networks and the resulting convex MPC schemes in experiments on a practical building application with the goal of minimizing energy consumption while maintaining occupant comfort, and in a numerical case study. The building has a water-based emission system and is located in temperate climate. 
We demonstrate that Predictive Control leads to savings between 26\% and 49\% of heating and cooling energy, compared to the building's baseline hysteresis controller. Moreover, we show that all model types lead to satisfactory control performance in terms of constraint satisfaction and energy reduction. 
However, we also see that the physics-informed ARMAX models have a lower computational burden, and a superior sample efficiency compared to the Machine Learning based models. Moreover, even if abundant training data is available, the ARMAX models have a significantly lower prediction error than the Machine Learning models, which indicates that the encoded physics-based prior of the former cannot independently be found by the latter.

\end{abstract}

\begin{keywords}
Building energy management \sep Data Predictive Control \sep Model Predictive Control \sep Physics-informed Machine Learning \sep Validation in experiment
\end{keywords}

\maketitle

\section{Introduction}
\label{Introduction}

\subsection{Motivation and contribution}

Buildings are responsible for approximately 40\% of the global final energy consumption \citep{IEABuildingEnergyPerformanceMetrics2015}, of which a large fraction is caused by space heating and cooling \citep{Drgona2020}. Besides retrofitting heating and cooling equipment and the envelope of a building, which is expensive, advanced control methods can be used to reduce energy consumption \citep{HameedShaikh2014}.
One of these methods is Model Predictive Control (MPC) \cite{Morari1999}. Here, a mathematical model of the system dynamics is used to optimize control inputs with respect to a cost function and comfort constraints in a receding prediction horizon.

MPC in buildings has been applied successfully in simulation \citep{Oldewurtel2012a,Touretzky2014,Chen2013,Mai2015} and real buildings \citep{Sturzenegger2016,Hilliard2017,Castilla2014,Ma2014,Zacekova2014a,Hammer2003} many times with significant reduction of energy consumption compared to the baseline controllers. However, models in building MPC are conventionally based on physics \citep{Picard2015,Sturzenegger2014}, which means that the model is built using principles of heat transfer and thermodynamics, or based on expensive excitation experiments, or a combination of both. Developing and maintaining such models is therefore often considered too expensive to justify investment, an issue that potentially hinders the commercial application of MPC in buildings \citep{Sturzenegger2016}.

As a result, data-driven approaches that rely purely on historical measurement data have emerged. Data-driven methods, which are sometimes referred to as Data Predictive Control (DPC) \citep{Smarra2018a}, either use models built from measurement data in an MPC framework \citep{Smarra2018a} or compute optimal inputs directly from past and currently measured data \citep{Coulson2019,Markovsky2006}. As the distinction between these methods and methods based on excitation experiments is blurry, we will use the common term MPC in the following for all predictive methods. Many Machine Learning (ML) methods, such as Artificial Neural Networks (ANN), Random Forests (RF) and Support Vector Machines (SVM) are universal function approximators \citep{Hornik1991,Hammer2003}, and have proven successful in various technical domains \citep{Silver2017,Wu2016,Hershey2017}. As they come with the tempting promise to model any system well as long as the underlying data quality is good, they are also natural candidates for MPC in buildings.

In \citep{Bunning2020} we successfully applied a combination of RF and linear regression, based on the work of \citep{Smarra2018a} to the cooling of a real apartment in Switzerland. In \cite{Bunning2020a} we have performed a similar study with models based on Input Convex Neural Networks (ICNN), which were extended from \cite{Amos2017} to ensure convexity in the face of recursive evaluation of the networks in an optimization framework. Under some assumptions on the cost function and state constraints, both RF and ICNN models lead to convex optimization problems that can be solved to global optimum in real time. In both studies, the controllers were able to reduce the energy consumption of the apartment while keeping room temperatures within comfort constraints during the vast majority of time. Other authors have conducted similar studies either in simulation or experiment, for an overview see Section \ref{subsec: Related work}. 

In the spirit of \cite{Mania2018}, it is, however, still unclear how these models compare to simpler identification methods, such as Autoregressive–Moving-Average with Exogenous Inputs (ARMAX) models identified through linear regression. While ARMAX models are not universal function approximators, they do have the trait of being physically interpretable, which is often not the case for ML methods. As we point out in Section \ref{subsec: Related work}, besides grey-box modeling, there is a significant gap in the literature for physics-informed data-driven models in the building domain. More identified gaps are the comparison of the performance of different data-driven models on real systems and a general evaluation of the benefits of data-driven MPC in practical building applications (see \citep{Kathirgamanathan2021} and Section \ref{subsec: Related work}). Methods that are applied to practical case studies often use data-driven modeling methods that lead to computationally complex non-convex optimization problems.

\begin{figure}
	\centering
		\includegraphics[width=0.45\textwidth]{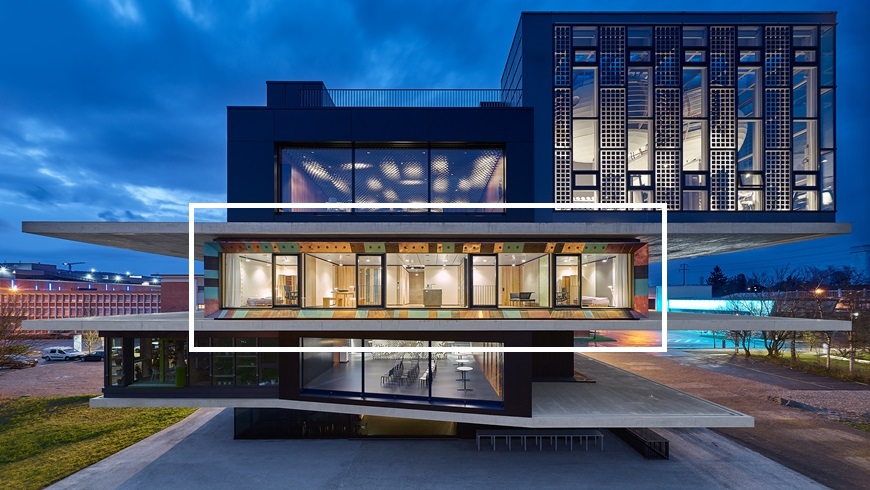}
	\caption{NEST building with the UMAR apartment unit marked in white, \textcopyright { }Zooey Braun, Stuttgart.}
	\label{fig: Nest}
\end{figure}

In this work, we exploit the physical interpretability of ARMAX models and their relation to the physical dynamics of buildings to create physics-informed ARMAX models. We conduct a series of heating and cooling experiments on an occupied apartment (Figure \ref{fig: Nest}), with the model embedded in a MPC framework with the goal of minimizing energy consumption while maintaining occupant comfort, and compare it to experiments conducted with RF and ICNN models in terms of constraint satisfaction and exploitation, and in terms of the computational burden, i.e the online optimization time and memory requirements. All presented modeling methods lead to convex MPC problems. Moreover, we generally evaluate the energy savings achieved with MPC, which amount to 26\% to 49\% compared to the baseline hysteresis controller used in the apartment. In a numerical experiment on the basis of historical measurement data from the apartment, we also compare the models in terms of sample efficiency, i.e. how much data a good model needs for training, and multi-step prediction accuracy. While all models qualitatively perform well on the task of predictive control in the application case study and outperform the baseline hysteresis controller in terms of energy consumption, the physics-informed ARMAX models show a considerably lower computational burden, better sample efficiency and better prediction accuracy in the numerical evaluation. On  one hand, our results show that any reasonably predictive model will be suitable for predictive control. On the other hand, they demonstrate that overly complicated ML-based models do not have any advantage over a physics-informed ARMAX model, i.e. they do not identify potential non-linearities such as time related solar gains or occupancy effects. On the contrary, the stronger physical prior of the ARMAX models, besides leading to higher sample efficiency, is not achieved by the ML models, even if abundant training data is available for training. These findings and the lower computational requirements in terms of optimization time and memory usage render the ARMAX approach superior for practical implementation.

The remainder of the article is structured as follows. In the second part of this Section, we discuss related work in more detail. In Section \ref{sec: Problem statement}, we introduce the concept of MPC. In Section \ref{sec: Methodology}, we briefly outline the earlier models based on RF and ICNN, and introduce the physics-informed ARMAX models. Section \ref{sec: Test bed} describes the apartment that we use as a test bed and its heating and cooling system. The results of the experiments of all models applied to the test case are presented and discussed in Section \ref{sec: Experiment case study}, while in Section \ref{sec: Numerical case study}, we investigate the sample efficiency of the models. Section \ref{sec: Conclusion} summarises our findings and provides directions for future work.

\subsection{Related work}
\label{subsec: Related work}

Several recent \citep{Maddalena2020,Kathirgamanathan2021,Drgona2020,Pean2019,Lee2020,Mariano-Hernandez2021,Gholamzadehmir2020,Zong2019} and less recent \citep{Rockett2017,Henze2013,Afram2014,Serale2018} reviews provide an excellent overview on the issues related to the use of MPC in buildings, including physics-based methods such as the Building Resistance-Capacitance Modeling (BRCM) toolbox \citep{Sturzenegger2014} or Modelica-based approaches \citep{Picard2015}. Here, we focus on the most relevant work related to our study, i.e. comparisons of different data-driven MPC models, practical applications of data-driven MPC, and physics-informed data-driven models in the building domain.

There are several studies on the comparison of data-driven MPC methods in simulation, such as comparing MPC controllers based on resistor capacitor (RC) models and feed-forward ANN in a simulation case study on a one-zone TRNSYS model \citep{Mugnini2020}. In the considered study, the RC models are based on knowledge of the thermal and geometrical building features, while the ANN model is purely data-driven. Both models show similar performance in terms of open-loop prediction. The RC model outperforms the ANN in closed-loop MPC in terms of comfort-constraint violation, while both controllers use significantly less energy than the baseline controller. However, both MPC controllers cause comfort constraint violations in more than 24\% of the operation time. There is no exact information on how the non-convex ANN MPC problem is solved.
In \citep{Wang2019a}, a variety of data-driven models based on Autoregressive with Exogenous Input (ARX), SVM, ANN and an RC model, applied to three different MPC simulation case studies in EnergyPlus, are compared. The models are trained on the basis of a single week of training data, generated with sinusoidal inputs signals. In closed-loop application, all models perform similarly well in the tasks of reference temperature tracking and energy minimization while respecting comfort constraints.
In another study, an MPC controller with a white box model is compared to one with an RC model parametrized with measurement data in a validated 12 zone simulation \citep{Picard2016}. While both models lead to efficient control behaviour as long as adequate data sets are available for system identification, the MPC controller based on the white box model consumes 50\% less energy. The authors stress that the results should be validated on a real system.
The authors of \citep{Ferkl2010} compare state space models obtained through subspace identification with ARMAX models on a nine-day historical data set from a real university building. They find that the subspace method outperforms the ARMAX model in terms of predictions on a validation data set. The models are not applied in closed-loop MPC.
In \cite{Chen2019}, models based on Recurrent Input Convex Neural Networks (RICNN), Recurrent (non-convex) Neural Networks (RNN) and RC models are compared in a simulation case study of an EnergyPlus office building. The training data length is ten months. Both Neural Network models show a better performance than the RC model in open-loop prediction. In closed-loop MPC, the RICNN consumes the least amount of energy, followed by RNN and RC respectively. The complexity and training method for the RC model is not discussed.

Other studies apply data-driven MPC for buildings in practice but do not compare different modelling methods. For example, the authors of \citep{Ferreira2012} use an ANN-based MPC controller on a set of rooms in a real university building. The controller saves up to 50\% of energy compared to the baseline controller.
Other authors use switched linear input-output models, which are trained based on excitation experiment data, to control a university building with an MPC framework \cite{Aswani2012}. In a eight day experiment, the controller saves a significant amount of energy compared to the baseline controller, while providing similar comfort levels.
In \citep{Finck2019}, an ANN-based MPC controller is applied with varying cost functions to a residential building. In multiple experiments of durations between four hours and one day, the authors show that the buildings' demand flexibility is maximized.
We have applied models based on RF and ICNN respectively with MPC in a real apartment building in Switzerland \cite{Bunning2020,Bunning2020a}. Both controllers are able to keep the room temperature between comfort bounds during the vast majority of time while minimizing energy consumption. The RF-based controller significantly reduces the cooling energy consumption compared to the baseline hysteresis-controller \cite{Bunning2020}. To the best of our knowledge, there is no comparison of different data-driven MPC methods on physical systems available in the literature. A review article \citep{Kathirgamanathan2021} considers both aspects, comparison on suitable data sets and practical validation of data-driven MPC controllers, as not appropriately addressed in the literature so far.

There is a general growth in research on physics-informed ML (see e.g. \citep{Raissi2017,Raissi2019,Sharma2018,Manek2020,Lutter2019,Marquez-Neila2017}), but very little work has been done so far in the domain of building energy control. Physics-informed ML differs from grey-box modeling \citep{Dimitriou2015,DeConinck2015,Li2021} in the sense that ML methods lead to input-output models, whereas in the building control domain, grey-box models are usually state-based models. Moreover, for grey-box models, the basic physical and architectural structure of the considered building is modeled by hand, which also requires a considerable amount of manual work, and parameters are fitted to measurement data.

In the field of physics-informed ML, the authors of \cite{Drgona2020a} introduce physics-constrained RNN to model the thermal dynamics of buildings and use information about the general model structure of buildings to structure the neural dynamics models, constrain the eigenvalues of the model, and use penalty methods to impose physically meaningful boundary conditions to the learned dynamics. The method is applied to an open-loop prediction on a data set of a real 20-zone building. The authors find that the prediction accuracy is significantly improved compared to not-constrained RNN models. The method is not tested in closed-loop MPC. 
In \citep{Zeng2021}, a method for simultaneous plant and disturbance identification to for buildings is presented. The authors use the Lasso method to promote sparsity and constrain coefficients to obtain linear time-invariant input–output stable models with positive DC gains. The presented method outperforms the grey-box method of \citep{Kim2016} in open-loop room temperature predictions in two data sets, one of which is simulated and the other one obtained from real building measurements. In \citep{Zeng2021a}, the approach is validated in simulation in a closed-loop experiment.
To the best of our knowledge these are the only published studies on the topic of physics-informed data-driven models in the building control domain. There is, to the best of our knowledge, no study available on closed-loop implementation on a physical building.

In addition to the lack of physics-informed data-driven models, the literature lacks comparisons of the performance of different data-driven models in MPC for practical building applications \citep{Kathirgamanathan2021}. Moreover, in cases where data-driven methods are indeed tested on real systems, the resulting MPC problem is often non-convex and has to be solved with non-linear solvers. While this approach might be suitable for single building energy management, the resulting computational complexity of solving the MPC problem online will likely lead to intractable problems if more complex system architectures or applications are considered. An example of such a problem is coordinated building control \citep{Lefebure2021}.

\section{Problem statement}
\label{sec: Problem statement}

In this study, we apply MPC to buildings with the aim of minimizing energy consumption while maintaining occupant comfort in terms of the room temperature. 
The buildings are equipped with water-based heating and cooling systems, which are present in most residential buildings and many legacy commercial buildings in central Europe. The system is actuated by manipulating the supply valves of the heating/cooling emission system.\footnote{The framework both supports continuous valves and on/off valves, as a continuous control signal can be translated to on/off commands through pulse-width modulation on a lower control level.} The available sensors are room temperature sensors and optionally measurements of the heating/cooling energy consumption of the entire building, and of the supply temperature. Moreover, we assume that weather forecasts for the ambient temperature and global solar irradiation are available.

MPC is a control scheme where a constrained optimization problem is solved repeatedly to find optimal control inputs over a receding horizon. Besides state-space formulations, the problem can be formulated with an input-output model, where previous outputs, inputs and disturbances are measured, and the optimization problem 

\begin{subequations}
\label{eqn:mpcmethod}
\begin{alignat}{2}
 \min_{\substack{u,y}}   &\quad && \sum_{k=0}^{N-1} J_k(y_{k+1},u_k)\label{eqn:mpctheory1}\\
\text{s.t. }   &\quad && \quad y_{k+1} = f(y_{k-},u_{k-},d_{k-})      \label{eqn:mpctheory2}\\
 &\quad && \quad (y_{k+1},u_k) \in (\mathcal{Y}_{k+1},\mathcal{U}_k)     \label{eqn:mpctheory3}\\
 &\quad && \forall k \in [0,...,N-1]     , \nonumber
\end{alignat}
\end{subequations}

\noindent is solved at discrete time instants. Here, the variables $y$, $u$ and $d$ denote the system outputs, control inputs and disturbances respectively. The variable $k$ denotes the time step in the horizon $N$, $J_k$ is the stage cost, and $f$ denotes the model describing the system dynamics as a function of autoregressive terms of outputs, and moving average terms of inputs and disturbances (or their forecasts), denoted by the subscript $-$. Output and input constraints are formulated with the sets $\mathcal{Y}_k $ and $\mathcal{U}_k$. 

Problem \eqref{eqn:mpcmethod} is solved every time a new measurement of the output is available; after each optimization, the first element of the optimal input sequence, $u_0^*$, is applied to the system and the process is repeated. Input-output models are suitable for building control, as there are usually no constraints on hidden system states (for example wall temperatures). 

Problem \eqref{eqn:mpcmethod} describes a basic MPC scheme, but many alternatives have been explored in the literature, for example applying multiple steps of the optimal sequence before repeating the optimisation, treating uncertainty in the disturbance forecast in a worst-case or stochastic way, using state space formulations for the system dynamics and state estimators during controller application, etc. The interested reader is referred to \citep{Kouvaritakis2016} for information on many of these alternatives.

The control performance generally improves with solving \eqref{eqn:mpcmethod} more frequently (i.e. the controller has a smaller sampling time) and a larger prediction horizon $N$. This, however, implies that less time is available to solve a more complex optimisation problem. As convex optimization problems can usually be solved more efficiently, formulations where the cost and constraints in \eqref{eqn:mpcmethod} are convex with respect to the decision variables are often preferable. Indeed, the art of MPC design is making design choices to master this trade-off between computation and control performance.

\section{Methodology}
\label{sec: Methodology}

In this section, we briefly describe models developed in our earlier work, based on RF and ICNN, then introduce the physics informed ARMAX models. 

\subsection{Random Forests with linear regression leaves}
\label{subsec: RF}

In \citep{Bunning2020} we presented a model, extending the work of \citep{Smarra2018a}, to describe the building dynamics in \eqref{eqn:mpctheory2} with a combination of RF and linear regression. The model is an input-output model $y=f({X}_c,{X}_d)$, where ${X}_c$ denotes the set of controllable inputs (for example valve positions, supply temperatures, etc.), and ${X}_d$ denotes the set of uncontrollable inputs, i.e. disturbances (for example ambient temperature, solar irradiation, time features, etc.) and previously measured room temperatures.

The model is built in two steps. First, a random forest $g({X}_d)$ is built, which maps the uncontrollable inputs ${X}_d$ to a finite set of leaves ${1,...,L}$. Second, in each leaf of the forest, a linear regression $h_i({X}_c)$ with $i \in {1,...,L}$ is performed on the basis of the controllable inputs, which maps $X_c$ to $y$. The resulting prediction function is $f(X_c,X_d)=h_{g({X}_d)}(X_c)$.

When the model is applied in MPC as \eqref{eqn:mpctheory2}, it is not applied recursively. Instead, for each timestep $k$ in the horizon $N$, a separate model $f_k$ is built. Here, only previously measured outputs, but not previously predicted outputs are used as model inputs to keep the optimization problem convex. For example, we would define $y_1=f_1 \left( X_c=(u_0,u_{-1}),X_d=(y_0,y_{-1},d_0,d_{-1}) \right)$ for the first prediction, but $y_2=f_2 \left( X_c=(u_1,u_0,u_{-1}),X_d=(y_0,d_1,d_{0}) \right)$ for the second prediction step; i.e. $f_2$ is not a function of $y_1$, only of $u_1$. The optimization problem is solved by looking up the leaves in the forest on the basis of measurements and forecasts of $X_d$ and collecting the appropriate functions $h_i({X}_c)$ first, and second by solving the resulting optimization problem. For a more detailed description of the approach, we refer to the original sources \citep{Bunning2020, Smarra2018a}.

\subsection{Input convex neural networks}
\label{subsec: ICNN}

In \citep{Bunning2020a}, we extended the work of \citep{Amos2017}, to describe building dynamics with a Neural Network that allows a convex formulation of problem \eqref{eqn:mpcmethod} in the absence of lower output constraints. The model is an input-output model $y=f(X_{cvx},X_{ncvx})$. The neural network architecture developed in \citep{Amos2017,Bunning2020a} ensures that the output $y$ is convex with respect to $X_{cvx}$ (for example decision variables such as room temperatures and control inputs), but not necessarily with respect to $X_{ncvx}$ (for example disturbances such as solar irradiation or ambient temperature). Furthermore, when the network is applied recursively, i.e. $y_1=f(X_{cvx,1},X_{ncvx,1})$ and $y_2=f(X_{cvx,2}=\{\tilde{X}_{cvx,2},y_1\},X_{ncvx,2})$, the output of the second time step $y_2$ is still convex with respect to the elements of the convex input of the first time step $X_{cvx,1}$. Here, $\tilde{X}_{cvx,2}$ denotes the convex inputs that are not related to the previous time step, i.e. the control input of step 2. If all model inputs are contained in $X_{cvx}$, then there is no need to include $X_{cnvx}$ in the formulation, and the model is referred to as a Fully Input Convex Neural Network (FICNN); otherwise, it is called a Partially Input Convex Neural Network (PICNN). With reference to the dynamics in \eqref{eqn:mpctheory2}, the outputs $y$ and inputs $u$ are assigned to $X_{cvx}$, and all disturbances $d$ can be assigned to $X_{ncvx}$; the model is therefore generally a PICNN.

Thanks to the structure imposed on the network, when applied in the MPC problem \eqref{eqn:mpcmethod}, constraint \eqref{eqn:mpctheory2} is convex non-decreasing with respect to the decision variables. It can then be shown that, if the input constraints in \eqref{eqn:mpctheory3} and the cost function \eqref{eqn:mpctheory1} are convex, problem \eqref{eqn:mpcmethod} is convex, as long as the outputs are box-constrained and no lower bounds are imposed. We note however, that in many practical cases, the problem could remain convex also in the presence of lower output constraints due to the monotonicity of the dynamics.

\subsection{Physics-informed ARMAX models}

The models presented in subsections \ref{subsec: RF} and \ref{subsec: ICNN} fall into the family of Machine Learning methods. In the case of modelling building dynamics, it is tempting to use them to learn the building's behaviour, as some dynamics-related effects are difficult to model from first principles for individual buildings. An example is the heating gain through windows as a function of time, global solar irradiation, window size, and window orientation. However, the ML methods usually do not encode constraints coming from building physics, for example the law that heat flows from warm to cold, the first law of thermodynamics, etc. In the following, we aim to construct a physics-informed data-driven model, which happens to be linear, as most dynamic effects on the thermal mass of a building are indeed linear.

\subsubsection{Modelling thermal zones}

The evolution of the temperature $T$ of a lumped mass in a building, for example a thermal zone, can be described by

\begin{equation}
m c_p \frac{dT}{dt} = \dot{Q}_{amb}+\dot{Q}_{n}+\dot{Q}_{sol}+\dot{Q}_{occ}+\dot{Q}_{act},
\label{eq: building dynamics}
\end{equation}

\noindent where $m$ and $c_p$ denote the mass and specific heat capacity of the mass respectively. The terms $\dot{Q}$ denote incoming and outgoing energy flows, i.e heat flows from and to ambient $\dot{Q}_{amb}$ and neighboring zones $\dot{Q}_{n}$, by solar irradiation $\dot{Q}_{sol}$, by occupancy $\dot{Q}_{occ}$ and by actuators $\dot{Q}_{act}$ (such as radiators, air conditioning, etc.) respectively. The terms $\dot{Q}_{amb}$ and $\dot{Q}_{n}$ are linear functions of $T$, for example 
$\dot{Q}_{amb}={\theta}_{amb}(T_{amb}-T)$, where ${\theta}_{amb}$ is a constant; $\dot{Q}_n$ is given by a sum of similar linear functions with $T_{amb}$ replaced by the temperatures of the neighbouring zones. The influence of occupancy $\dot{Q}_{occ}$ is part of ongoing research and is neglected for this particular model, partly because occupancy forecasts were not being available in the case study. We will treat the remaining terms, $\dot{Q}_{sol}$ and $\dot{Q}_{act}$, in the following.

\subsubsection{Modeling solar gains $\dot{Q}_{sol}$}

A simple physics-based model for solar gains through windows is given by

\begin{equation}
\dot{Q}_{sol}=A_{win}\sin(\alpha-{\alpha}_0) \frac{\cos(\beta)}{\sin(\beta)} I_{hor},
\label{eq: solar1}
\end{equation}

\noindent where $A_{win}$ is the window surface area, and the angles $\alpha$ and $\beta$ denote the azimuth (i.e. the horizontal angle with respect to north) and elevation (i.e. the vertical angle with respect to earth's surface) of the sun respectively. The offset $\alpha_0$ denotes the orientation of the window and $I_{hor}$ is the horizontal global irradiation, which is an input commonly available from a weather forecast. Note that $sin(\alpha-{\alpha}_0)$ can become negative, which means that the direction of the irradiation through the surface is negative. For the given case of a window, we therefore take $\max(0, sin(\alpha-{\alpha}_0))$. Equation \eqref{eq: solar1} shows that the gains through solar irradiation are a non-linear function of the  window orientation and the angles $\alpha$ and $\beta$, which are themselves highly non-linear functions of time. 

To embed $\dot{Q}_{sol}$ in our model, we assume that the second term of \eqref{eq: solar1}, $I_{vert(t)}=\frac{\cos(\beta)}{\sin(\beta)} I_{hor}$, which denotes the irradiation on a vertical surface following the sun, is given as an input. This is a reasonable assumption as $I_{hor}$ can be obtained from a weather forecast and $\beta$ is a function of time and location. We model the first term, $A_{win}\sin(\alpha-{\alpha}_0)$, with $\tau$ time varying coefficients $[\theta_{sol,1} \; ... \; \theta_{sol,\tau}]$, where $\tau$ represents one day:

\begin{equation}
\dot{Q}_{sol}=[\theta_{sol,1} \; ... \; \theta_{sol,\tau}] [I_{vert,1} \; ... \; I_{vert,\tau}]^T ,
\label{eq: solar onehot}
\end{equation}

\noindent where $[I_{vert,1} \; ... \; I_{vert,\tau}]$ is a one-hot encoding \cite{Brownlee} of $I_{vert}$ with respect to discrete time-periods $1,...,\tau$:

\begin{equation}
I_{vert,i}= 
\begin{cases}
I_{vert(t)},& \text{if } i\geq t > {i+1}\\
0, & \text{otherwise.}
\end{cases}
\label{eq: solar2}
\end{equation}

This way of modelling the solar irradiation creates $\tau$ input variables that are zero during most times of the day, but are equal to $I_{vert(t)}$ for fixed periods. For example, in the case of $\tau$=4 with time periods of equivalent length, $I_{vert,1}$ attains $I_{vert(t)}$ for the first six hours of the day, and is zero for all other times, $I_{vert,2}$ attains $I_{vert(t)}$ for the period of 6 am to 12 pm, and zero otherwise, etc. The solar gains $\dot{Q}_{sol}$ are now a linear function of the easy-to-obtain inputs $I_{vert,t1} \; ... \; I_{vert,\tau}$. A validation of this modelling approach compared to the physical model of \eqref{eq: solar1}, on the data set later used in the case studies, is shown in Figure \ref{fig: Sol valid} in the Appendix for $\tau = 9$. The coefficient of determination $R^2$ is 0.96 for both the training and the testing set. Note that the number of one-hot encoded inputs $\tau$ does not relate to the sampling time of the MPC; it only needs to be sufficiently high to reach a reasonable fit of $\dot{Q}_{sol}$.

\subsubsection{Modeling actuator gains $\dot{Q}_{act}$}
\label{subsec: actuator gains}

For most relevant heating and cooling systems, such as radiators, floor heating and (neglecting the water and vapour content) air conditioning units, the energy transferred from an actuator to a thermal zone can be described by an equation of the form

\begin{equation}
\dot{Q}_{act}=\dot{m}_f c_{p,f} (T_{sup}-T_{ret}). 
\end{equation}

\noindent Here, $\dot{m}_f$ and $c_{p,f}$ denote respectively the mass flow and specific heat capacity of the fluid (usually water or air), and $T_{sup}$ and $T_{ret}$ denote supply and return temperatures.

There are a variety of options to model $\dot{Q}_{act}$. In an ideal case, it is directly accessible by measuring $\dot{m}$, $T_{sup}$ and $T_{ret}$ and can be used directly in \eqref{eq: building dynamics}. Unfortunately in many buildings measurements at this level are not available. In such cases, one has several options for inferring Q from the available measurements. One option is by measuring the total energy consumption of a building and allocating portions of it based on design mass flows to individual rooms; this is the approach we follow for $\tilde{\dot{Q}}_{act,i}$ in Section \ref{sec: Test bed}. Another is to model $\dot{Q}_{act}$ as a linear function of the mass flow or a valve position $b$:

\begin{equation}
\dot{Q}_{act}= \bar{\theta} \dot{m} \approx \theta b.
\label{eq: act2}
\end{equation}

\noindent This assumes that $(T_{sup}-T_{ret})$ is approximately constant, which is a reasonable assumption for example in the case of heating with a high supply temperature. In the case of using the valve position $b$ and not directly the mass flow $\dot{m}$, the assumption that their relation is linear also has to be made.

A last option is suitable when both valve position $b$ and $T_{sup}$ are measured. The transferred energy then follows

\begin{equation}
\dot{Q}_{act}= \theta b(T_{sup}-{T}),
\label{eq: act3}
\end{equation}

\noindent where ${T}$ is the current temperature of the zone itself\footnote{The input is therefore linear in the system state.}. For air-based systems, the assumption $T_{ret}={T}$ is always reasonable. For water-based systems, the assumption holds if the heat transfer surfaces are large and the mass flows are low. Also, the assumption that $\dot{m}$ and $b$ have a linear relation has to hold again. All options assume that $c_{p,f}$ is constant. In the case studies in Sections \ref{sec: Experiment case study} and \ref{sec: Numerical case study}, we will explore several of these options for $\dot{Q}_{act}$.

\subsubsection{Positivity constraints for building dynamics}

By substituting all $\dot{Q}$ terms in \eqref{eq: building dynamics} using \eqref{eq: act3} for $\dot{Q}_{act}$, and neglecting occupancy, the dynamics can be rewritten as

\begin{equation}
\begin{aligned}
m c_p \frac{dT}{dt} = & {\theta}_{amb} (T_{amb}-T) + {\theta}_{n} (T_{n}-T) \\
& + \left( \sum_{t_i=t_1}^{\tau} {\theta}_{sol,t_i} I_{vert,t_i} \right) + {\theta}_{act} b(T_{sup}-\bar{T}),
\end{aligned}
\label{eq: building dynamics2}
\end{equation}

\noindent where we assume a single neighbouring zone to simplify the notation; similar equations are obtained with the other options for modelling $\dot{Q}_{act}$. To avoid the bilinearity in the valve opening $b$ and the zone temperature $T$, the zone temperature is replaced by an approximation $\bar{T}$, which can be obtained from the last measured room temperature for example. After performing Euler discretization, the discrete time thermal zone dynamics can be written as

\begin{equation}
\begin{aligned}
T_{k+1} = & \left(1-\frac{\Delta t \: {\theta}_{amb}}{m c_p} - \frac{\Delta t  \: {\theta}_{n}}{m c_p} \right) T_k \\
& + \left(\frac{\Delta t \: {\theta}_{amb}}{m c_p} \right) T_{amb,k} + \left(\frac{\Delta t \: {\theta}_{n}}{m c_p} \right) T_{n,k} \\
& + \left( \sum_{t_i=t_1}^{\tau} \left(\frac{\Delta t \: {\theta}_{sol,t_i}}{m c_p} \right) I_{vert,t_i} \right) \\
& + \left(\frac{\Delta t \: {\theta}_{act}}{m c_p} \right) b(T_{sup}-\bar{T})_k,
\end{aligned}
\label{eq: building dynamics3}
\end{equation}

\noindent where the subscripts $k$ and $k+1$ denote the current and subsequent time step respectively, and $\Delta t$ is the sampling time. By reducing all constants in \eqref{eq: building dynamics3} to coefficients $\tilde{\theta}$, the expression can be further simplified to

\begin{equation}
\begin{aligned}
T_{k+1} = & \tilde{\theta}_{auto} T_{k} + \tilde{\theta}_{amb} T_{amb,k} + \tilde{\theta}_{n} T_{n,k} \\
& + \left( \sum_{t_i=t_1}^{\tau} \tilde{\theta}_{sol,t_i} I_{vert,t_i} \right) + \tilde{\theta}_{act} b(T_{sup}-\bar{T})_k.
\end{aligned}
\label{eq: building dynamics4}
\end{equation}

\noindent We note that for a sufficiently small time step $\Delta t$, all coefficients $\tilde{\theta}$ are positive since the masses, specific heat capacities and heat transfer coefficients are all positive.\footnote{Here, conventional sampling times for building MPC, for example 15 or 30 minutes, are sufficient to keep $\tilde{\theta}_{auto}$ positive.}

The temperatures can be stacked in a state vector $x$, inputs ($b(T_{sup}-\bar{T})$) and disturbances ($T_{amb}$, $I_{vert,t_i}$) can be stacked in the input and disturbance vectors $u$ and $d$, and a conventional linear state space system of the form $x_{k+1} = \text{A} x_k + \text{B}_u u_k + \text{B}_d d_k$, $y_{k+1} = \text{C} x_k$ can be formulated, where the output vector y collects all the entries of $x$ (zone temperatures) that can be measured. The influence of the hidden states can then be modelled implicitly by a convolution of the previous outputs $y$ (i.e. the measured zone temperatures), inputs and disturbances. This results in an ARMAX model of the form

\begin{equation}
y_{k+1}=\Theta \: [y_k \: ... \: y_{k-\delta} \;  u_k \: ... \: u_{k-\delta} \; d_k \: ... \: d_{k-\delta}]^T,
\label{eq: ARX2}
\end{equation}

\noindent which can be used for the dynamics in \eqref{eqn:mpctheory2}. Here $\Theta$ is the vector of regression coefficients for each thermal zone and $\delta$ determines the number of autoregressive and moving average steps.\footnote{Generally, $\delta$ can be different for $y$ and $u$, $d$.} Similar to the state space system, the elements of $\Theta$ are positive and can be found through non-negative least squares regression.

\subsection{Model properties}
\label{subsec: Model properties}

The presented models have different properties in terms of expressiveness and the complexity of the resulting MPC problem. The ARMAX model is limited to \eqref{eqn:mpctheory2} being linear in all variables. RF are generally more expressive, as they are piecewise constant in the disturbances and piecewise linear in the inputs. ICNN are also more expressive than ARMAX, being convex in the inputs, and even less restrictive for disturbances. Generally it could be expected that more expressive models lead to a more accurate representation of the system dynamics as long as they can be trained to optimality and there is enough data to avoid overfitting. This is studied in Section \ref{sec: Numerical case study} below.

Compared to ICNN, ARMAX and RF are generally more lightweight in terms of online computation, when applied in the MPC problem \eqref{eqn:mpcmethod}. For example, if \eqref{eqn:mpctheory1} is quadratic and \eqref{eqn:mpctheory2} linear in the decision variables, the resulting problem is a QP, for which very efficient solvers are available. By contrast, ICNN lead to a general convex optimisation problem, that require more generic methods like interior point or gradient decent. To get a convex problem, ICNN are limited to box constraints with only an upper bound when it comes to the output variables in \eqref{eqn:mpctheory3}. The other two models allow general convex output constraints.

\section{Testbed}
\label{sec: Test bed}

\begin{figure}
\centering
\includegraphics[width=0.49\textwidth]{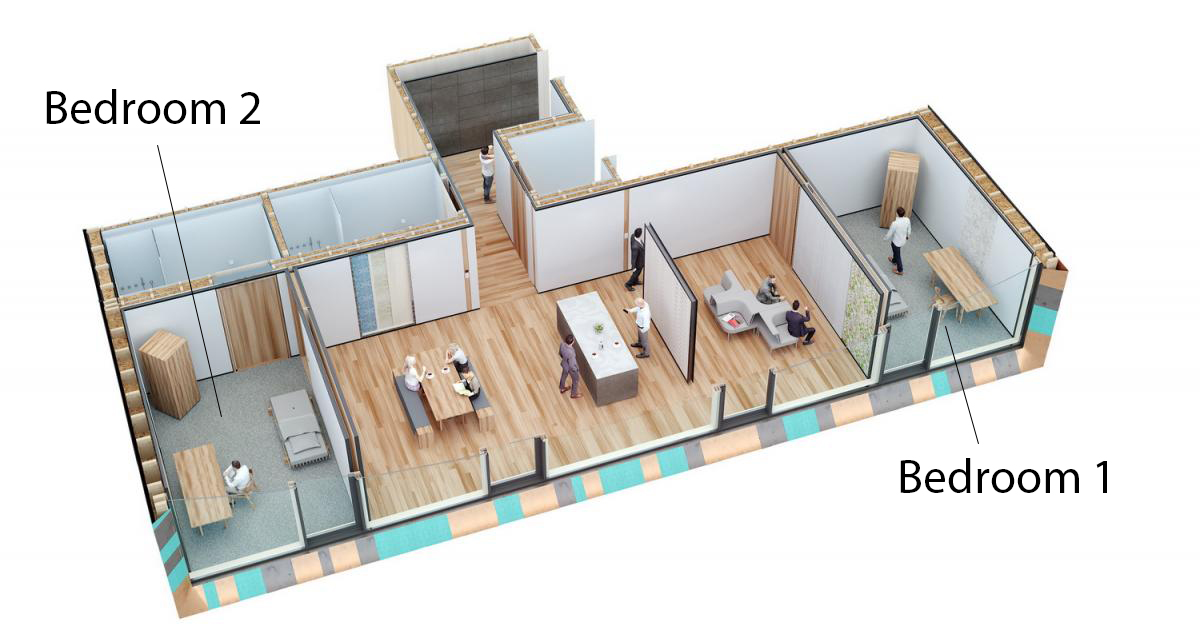}
\caption{Rendering of the UMAR unit in NEST with both bedrooms marked. \textcopyright {} Werner Sobek.}
\label{fig: UMAR rendering}
\end{figure}

We use the Urban Mining and Recycling (UMAR) unit \citep{Kakkos2019,Heisel2019} of the NEST building \citep{Richner2018} at Empa, Dübendorf, Switzerland as a test bed in our case studies. The unit, shown in Figures \ref{fig: Nest} and \ref{fig: UMAR rendering}, is an apartment built to demonstrate the circular economy in the building construction industry and is constructed from recycled material or material that can be recycled completely after dismantling the unit. It comprises two bedrooms and one living/kitchen area with large south-east facing windows, two bathrooms, an entrance area and a technical room. The two bedrooms have identical floor plans and furniture. The apartment is considered to be a ``living lab'' and is occupied by two persons, who live there.

\begin{figure}
\centering
\includegraphics[width=0.40\textwidth]{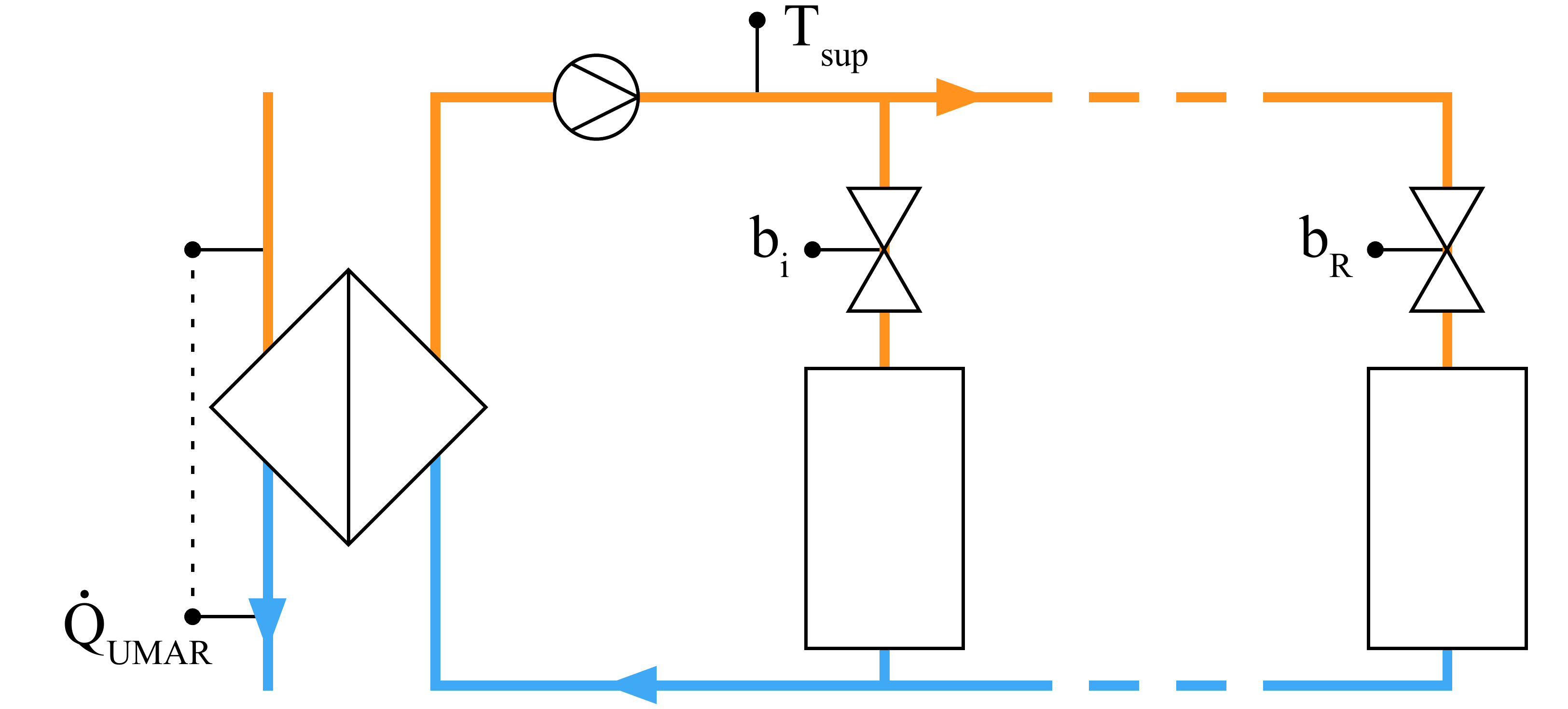}
\caption{Heating system of the UMAR unit in NEST. The heating loop is connected to the central heating system via a heat exchanger. The heating and cooling panels in each room are controlled with individual on/off valves.}
\label{fig: UMAR heating}
\end{figure}

The unit is equipped with a water-based heating and cooling system (Figure \ref{fig: UMAR heating}). The system is connected to the NEST heating grid via a heat exchanger and the total energy consumption of the unit, $\dot{Q}_{\text{UMAR}}$, is measured with the help of an Energy Valve\texttrademark \citep{Belimo2021}. Each room is equipped with at least one heating/cooling ceiling panel, which is controlled by individual on/off valves $b_i$. There is a central pump that provides pressure, as long as one of the valves is open. The supply temperature $T_{sup}$ is regulated by adjusting the mass flow on the NEST heating grid side of the heat exchanger. The same system is used for cooling during warm days through a second heat exchanger connected to the NEST cooling grid. As the pump delivers a constant pressure and the valves are either fully open or closed, the design mass flows\footnote{We introduce this way of calculating the transferred energy in Section \ref{sec: Test bed} rather than \ref{sec: Methodology}, and mark the variable with a tilde, as the design mass flows of each individual zone are commonly not known in a building and specific to this case study. They are specified by the heating/cooling system installer.} for each heating/cooling panel $\dot{m}_i$ can be used to calculate the amount of energy transferred to each room $\tilde{\dot{Q}}_{act,i}$ through

\begin{equation}
\tilde{\dot{Q}}_{act,i}=\dot{Q}_{\text{UMAR}} \frac{\dot{m}_i b_i}{\sum_{j \in \mathcal{R}} \dot{m}_j b_j},
\label{eq: Qact true}
\end{equation}

\noindent where $\mathcal{R}$ denotes the set of all rooms in the unit. All measurements, including the individual room temperatures are stored in an SQL database with a sampling time of one minute. A weather forecast, sampled in one-hour periods, for the ambient temperature and global solar irradiation is available from the national weather service, Meteo Swiss. All actuators can be controlled via OPC-UA software clients \citep{Leitner2006}. During standard operation, the room temperature is controlled by a hysteresis baseline controller that regulates the room temperature between an occupant-decided set point and 1 \degree C below. An example of a temperature trajectory of bedroom 2 and the corresponding control input with the baseline controller is shown in Figure \ref{fig: Baseline} in the Appendix.

\section{Experiment case study}
\label{sec: Experiment case study}

Over the course of two years, we have implemented MPC, based on RF, ICNN and physics-informed ARMAX models, for 156 full days of experiments in the two bedrooms of the UMAR unit. In the following, we will discuss the model configuration and training, present example closed-loop trajectories of the controllers, and analyse the energy consumption compared to the base-line controller. For each controller type a different model was trained for each bedroom.

\subsection{Model and controller configuration}
All models use the same training data set of 2018-05-23 to 2019-05-28 (370 days), generated during normal operation of the building with the baseline hysteresis controller. In this subsection, we define which inputs and outputs each model uses, how the model hyperparameters were determined, and how many degrees of freedom each model has.

To configure the RF model, extensive feature engineering and hyperparameter tuning was conducted \cite{Bunning2020}. For the feature engineering, domain knowledge was used to pre-select certain model inputs and disregard others. With these features, models were then trained on the first 70\% of the data and tested on the remaining data. As a result of feature engineering, it was found that predicting room temperature differences $\Delta x$ (or $\Delta T$), i.e. the change of room temperature between two sampling steps, leads to better model accuracy than directly predicting room temperatures. This has no effect on the convexity of the MPC problem \eqref{eqn:mpcmethod}. As model inputs for the forests, autoregressive terms of $\Delta x$, of temperature differences between rooms and neighboring rooms, of window opening times, of the horizontal solar irradiation, and of the ambient temperature were chosen. We also use the time of the day and month, encoded as cosine and sine functions\footnote{Both cosine and sine are used, to have distinct inputs for each time, as sine and cosine functions have the same function value twice per period \citep{Bescond2020}. We can also fit any phase offset with a linear parametrization.}, to capture any time-related influences on the room temperatures (for example, occupants entering the rooms every day at the same time) and to identify relations between time and global solar irradiation (similar to \eqref{eq: solar onehot}). For the linear regression in the forest leaves, moving average terms of the control input $\tilde{\dot{Q}}_{act,i}$ are used. The sampling time was set to 30 minutes based on preliminary numerical studies.

The hyperparameters of a random forest are mainly the number of trees per forest and the minimum number of samples per leaf. To find suitable hyperparameters, we applied line searching, where each parameter is varied while keeping the others constant. The procedure is not iterative, i.e. each parameter is just updated once. The resulting model has 200 trees per forest and a minimum of 200 samples per leaf.

The degrees of freedom of a random forest with linear regression grows with the number of training samples $S$, due to the automatic scaling through the minimum number of samples per leaf, and approximately follows $3 \frac{S}{200}$ (with 200 being the number of samples per leaf). For a sampling time of 30 minutes, this amounts to 270 for the full training set. The models are implemented with \textit{Scikit-learn} \citep{Pedregosa2011} in \textit{Python 3}.

The configuration of the ICNN models is described in \citep{Bunning2020a} and in more detail in \cite{Schalbetter2020}. For feature engineering, a k-fold cross validation with k=12 was performed on the entire data set, with 9 folds being used for training and 3 for validation. After this, the networks were chosen to predict the change of room temperatures instead of directly predicting temperatures. For PICNN, as non-convex inputs, autoregressive terms of the solar irradiation, the temperature difference with respect to neighboring rooms and the sine-encoded time of day (but not time of year) were chosen. As convex inputs, autoregressive terms of the change of room temperature, the temperature difference between room and ambient, and the heating/cooling control input were chosen. In the case of FICNN, all the above features are convex inputs. The sampling time was also decided on with cross validation and is 20 minutes for predictions up to one hour and 180 minutes beyond \cite{Schalbetter2020}.

The hyperparameters of an ICNN are the training method, the step-size of the training method, number of training epochs, nodes per layer, number of layers, and an offset in the ReLu activation functions. The hyperparameters were optimized with line searches applied to the k-fold cross validation. The resulting networks have approximately 1000 degrees of freedom (i.e. parameters to fit). The models are implemented with \textit{Keras} \citep{Chollet2018} in \textit{Python 3}.

The inputs for the ARMAX model directly follow from the physical structure described in Section \ref{sec: Methodology}. They comprise autoregressive terms for the room temperature\footnote{For linear models, the choices of using absolute temperatures or temperature differences both lead to the same model accuracy.}, moving average terms for neighboring zones, the ambient temperature, the one-hot encoded global solar irradiation and for the control input. The only hyperparameters to be tuned are $\tau$, the number of one-hot inputs of the solar irradiation, and $\delta$, the number of autoregressive terms. These were chosen to be $\tau = 9$ and $\delta = 3$ after preliminary experiments. Our ARMAX models have $(\delta+1)(4+\tau)$ degrees of freedom, i.e. 52 for our configuration. The models are implemented with \textit{Scikit-learn} \citep{Pedregosa2011} in \textit{Python 3}.

All models are applied as the model for the building dynamics \eqref{eqn:mpctheory2} in the MPC optimization problem \eqref{eqn:mpcmethod}. After specifying the cost function and constraints, this leads to the optimization problem

\begin{subequations}
\label{eqn:mpcmethodnew2}
\begin{alignat}{2}
\min_{\substack{u,y,\epsilon}}   &\quad && \sum_{k=0}^{N-1} (u_k R u_k + \lambda \epsilon_{k+1}) \label{eqn:mpctheorynew1}\\
\text{s.t. }   &\quad && \quad y_{k+1} = f({y}_{k-}, {u}_{k-},{d}_{k-})      \label{eqn:mpctheorynew2}\\
 &\quad && \quad y_{min} - \epsilon_{k+1} \leq y_{k+1} \leq y_{max} + \epsilon_{k+1}\label{eqn:mpctheorynew3}\\
 &\quad && \quad \epsilon_{k+1} \geq 0\label{eqn:mpctheorynew4}\\
 &\quad && \quad u_{min} \leq u_{k} \leq u_{max}\label{eqn:mpctheorynew5}\\
 &\quad && \forall k \in [0,...,N-1]     , \nonumber
\end{alignat}
\end{subequations}

\noindent where $\epsilon$ is a slack variable introduced to ensure feasibility. A quadratic weight for the control input $R$ and a linear weight $\lambda$ for the comfort slack variable are used in the cost function; the weights and prediction horizon are specified below for each experiment. A quadratic cost function is chosen because, based on preliminary experiments \citep{Schalbetter2020,Huber2019}, it provides a good balance between minimizing energy consumption and avoiding peaks, and yields a smoother control input trajectory compared to a linear cost function. The comfort constraints $y_{min}$ and $y_{max}$ are time varying and will also be reported with the results. The control input is applied to the valves of the UMAR unit with pulse-width modulation (PWM). The limits for the control input are therefore $u_{min}=0$ and $u_{max}=1$. The resulting problem is a Quadratic Program in the case of the ARMAX and RF models (see discussion in Sections \ref{subsec: RF} and  \ref{subsec: Model properties}), which we solve with the QP solver of \textit{CVXOPT} \cite{Andersen} in \textit{Python 3}. For the ICNN, the problem results in a convex problem without direct access to the function derivatives. We solve it with the \textit{COBYLA} \cite{Powell1994} solver of \textit{SciPy} \cite{Virtanen2020} in \textit{Python 3}.

\subsection{Example closed-loop experiments}
\label{subsec: trajectories}

\begin{figure*}
\centering
\includegraphics[width=1\textwidth]{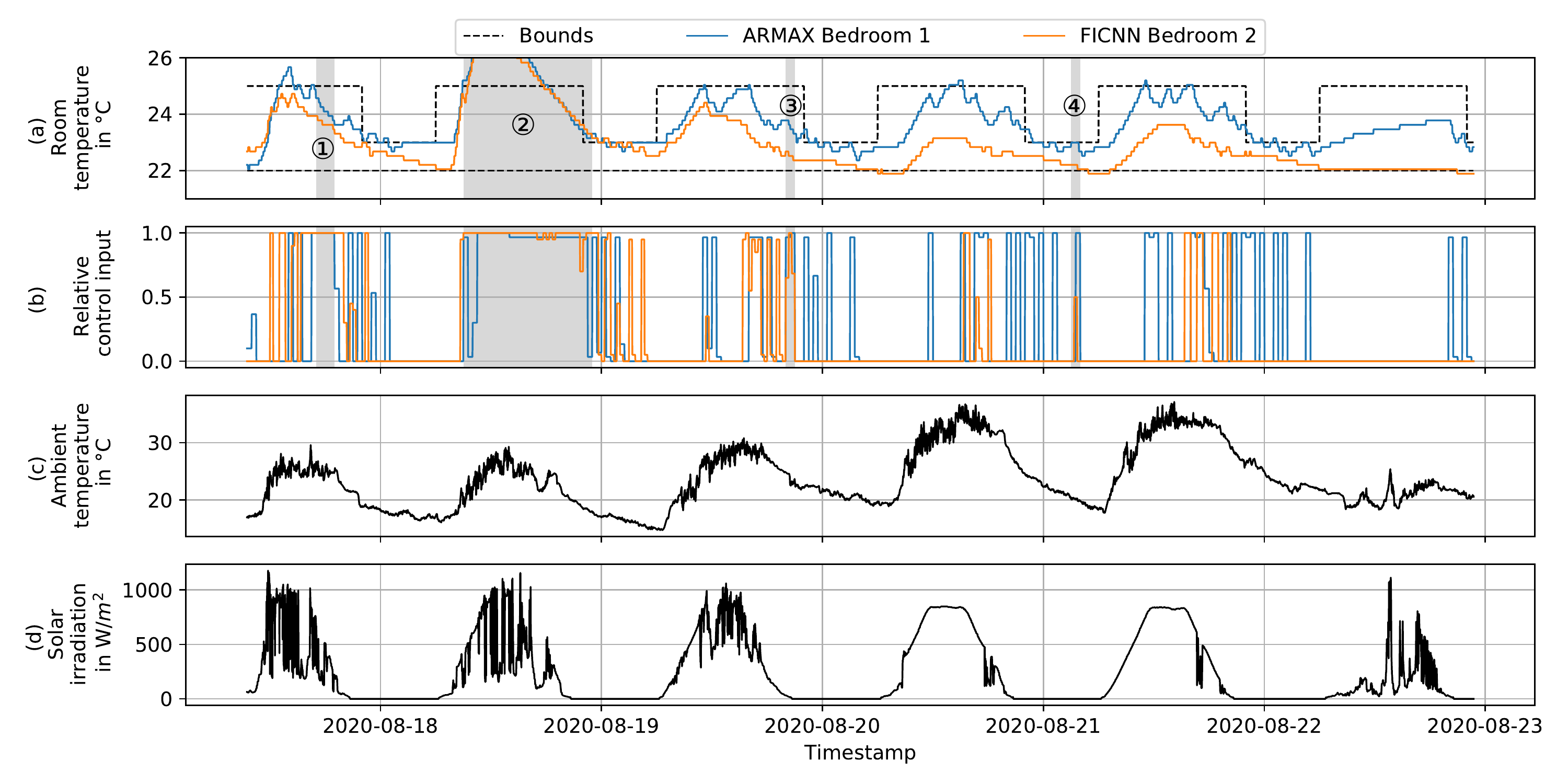}
\caption{Cooling experiment with ARMAX in bedroom 1 (blue) and FICNN in bedroom 2 (orange). (a): Temperature in the two bedrooms. The comfort bounds are shown in dashed black. In \protect\circled{1}, \protect\circled{3} and \protect\circled{4} the connection between the controller and actuators was lost for a short time. In \protect\circled{2}, the otherwise closed window blinds were automatically opened due to strong wind, which lead to the system not being able to reject the solar gains, even at maximum cooling power. (b): Relative control input, i.e. the fraction of time where the maximum control input is applied during one control step. (c): Measured ambient temperature at the experiment site. (d): Global solar irradiation at the experiment site.}
\label{fig: Cooling ARX FICNN}
\end{figure*}

\begin{figure*}
\centering
\includegraphics[width=1\textwidth]{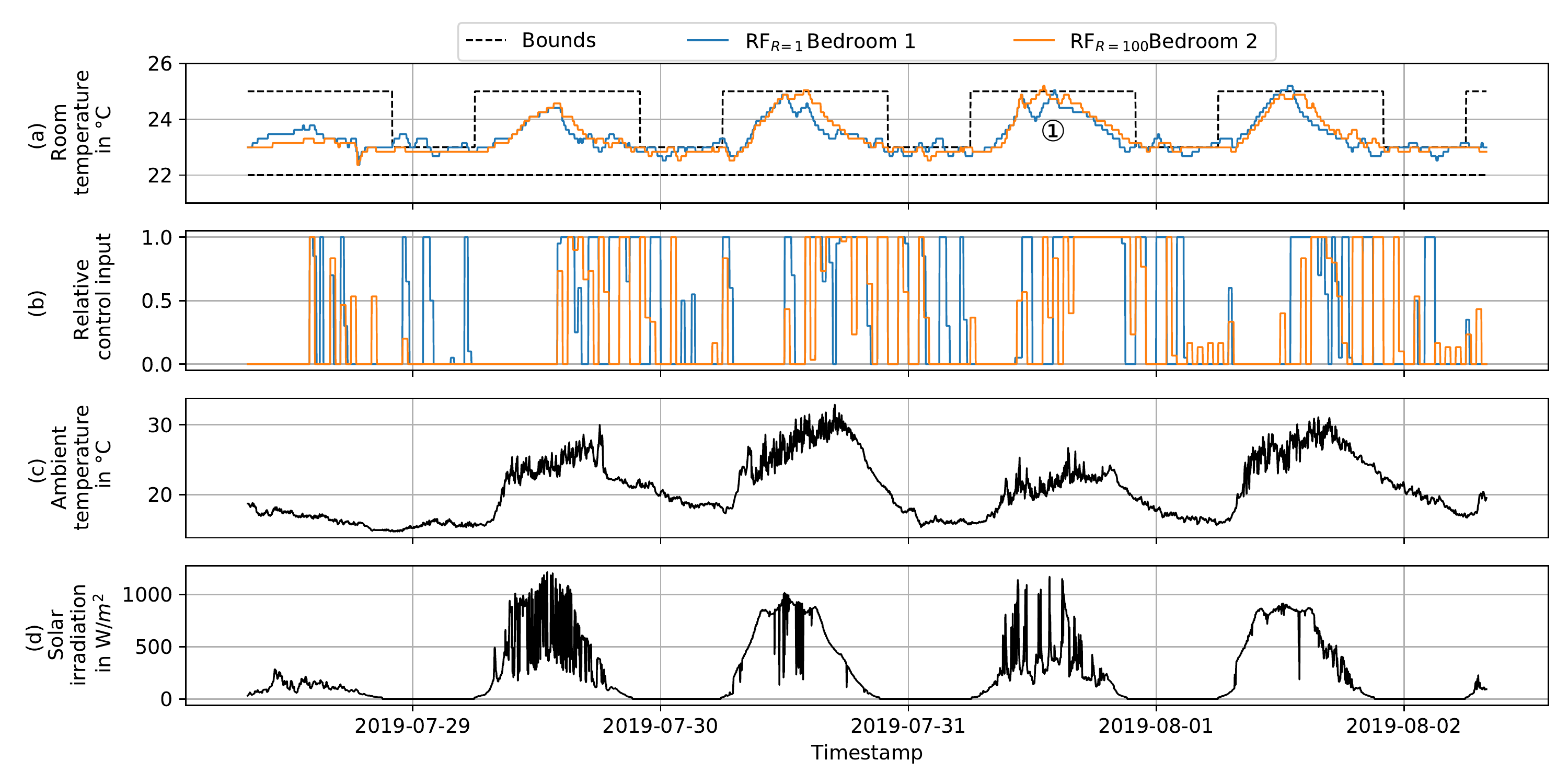}
\caption{Cooling experiment with RF in bedroom 1 (blue) and bedroom 2 (orange) with different weights on the control input cost. (a): Temperature in the two bedrooms. The comfort bounds are shown in dashed black. (b): Relative control input, i.e. the fraction of time where the maximum control input is applied during one control step. (c): Measured ambient temperature at the experiment site. (d): Global solar irradiation at the experiment site.}
\label{fig: Cooling RF}
\end{figure*}

To compare the closed-loop behaviour of the ARMAX controller and an FICNN controller, we conducted a cooling  experiment on NEST with the ARMAX model (controller properties: N=7h, R=1, $\lambda$=100, $\delta$=7, $\Delta t$=30 min, actuation: valve opening) applied to bedroom 1 and a FICNN (controller properties: N=7h, R=1, $\lambda$=100) applied to bedroom 2. Note that the relatively short horizon for a building application is sufficient in the given case because the considered apartment is light-weight and the heating/cooling emission system is fast. The controller does not benefit from a longer prediction horizon, as preliminary experiments have shown \citep{Huber2019}. The results in Figure \ref{fig: Cooling ARX FICNN} show that both controllers keep the temperature within the comfort constraints most of the time and exploit the relaxed constraints during the day to save energy. However, the controller using the ARMAX model is less conservative. It keeps the room temperature closer to the upper comfort constraint during the night, and meets the lowered comfort constraint at 22:00 just in time or violates it by a fraction of a degree for a short period of time. Day 2020-08-19 is characteristic of the different behaviour of the two controllers. Here, the FICNN-controller applies control action during the second half of the day although the temperature is already relatively low, (which even leads to slight violations of the lower comfort constraints a few hours later,) while the ARMAX-controller slightly violates the upper comfort constraint. 
Experiments with the PICNN showed similar results to those conducted with the FICNN (Data omitted in the interest of space). 

To compare the behaviour of the RF controller in a similar setting\footnote{In general, building MPC experiments are not reproducible due to changing (and uncontrollable) environmental conditions. Since bedrooms 1 and 2 are nearly identical architecturally, this experimental configuration represents the most feasible way to compare two controllers with similar environmental conditions.}, we performed a cooling experiment with an RF controller applied to bedroom 1 (controller properties: N=6h, R=1, $\lambda$=100); to investigate the effect of the cost weights we also applied the same controller to  bedroom 2 with controller properties N=6h, R=100, $\lambda$=100. Figure \ref{fig: Cooling RF} demonstrates that the behaviour is very similar to the one observed with the ARMAX model. The room temperature is kept close to the upper comfort constraint during the night, the controller exploits the relaxed comfort constraints during the day to save energy, and starts cooling early enough to meet the lowered comfort constraints at 22:00. The difference in relative weighting between costs for control input and constraint violations does not seem to significantly influence behaviour. 

In general, the experiments demonstrate that all controllers have reasonable behaviour, with the ICNN-based controllers being more conservative compared to the ARMAX and RF-based controller. This is possibly due to underestimating the influence of the control input or overestimating the thermal capacity of the system. Such effects are discussed in detail in \citep{Blum2019}. Although this issue is more pronounced for the ICNN, it is also visible for RF and ARMAX to a lesser extent during times where the comfort constraints are relaxed but the controller is aiming for the lowered upper comfort constraint at 22.00, for example at \circled{1} in Figure \ref{fig: Cooling RF}. Here, the controllers often apply a high control input in one step, which cools down the room temperature more than necessary, and then do not apply any control input in the consecutive time step, which lets the room temperature rise again. This is likely a result of the training data being correlated due to the underlying feedback controller, where the effects of control input and disturbance on the room temperature cancel out. For example in the heating case, cold ambient temperatures lead to high heating power, warm ambient temperatures lead to low heating power. For a well-regulated system, the regression is ill-posed because the underlying data is not \textit{persistently exciting} \citep{ljung1994global}.  Besides the obvious solution of generating training data with an uncorrelated input signal, tracking the predicted temperature trajectory with a lower-level feedback controller instead of directly applying the optimized control input could mitigate the issue.

Similar experiments in UMAR have been conducted for the heating case with the RF and the ARMAX controller. (We have not conducted heating experiments with the ICNN models due to the MPC problem not being convex at the lower comfort constraint \citep{Bunning2020a}.) The general behaviour of the controllers in the heating experiments is similar to the one observed in the cooling experiments. The results therefore do not add anything new to the discussion, other than the observation that predictive control in general also works in the heating case with both model types. We therefore refer the interested reader to the linked data repository for data on these experiments, see Section \textit{Data Availability}.

\subsection{Energy consumption}

We compare the energy consumption of the MPC-based controllers to the baseline controller on the basis of the concept of Heating Degree Solar Days (HDSD) and Cooling Degree Solar Days (CDSD), explained in the following.

\subsubsection{HDSD and CDSD}

Heating Degree Days (without \textit{Solar}) are commonly used to quantify the energy consumption of a building as a function of the ambient conditions \citep{ScienceDirect}. Here, a base temperature $T_{amb,b}$ is defined (usually assumed to be the ambient temperature at which no heating is necessary), and it is assumed that the daily heating energy consumption of a building is proportional to the difference between the daily average ambient temperature $\bar{T}_{amb}$ and the base temperature:

\begin{equation}
Q_{hea}=\theta_{\textit{HDD}}(T_{amb,b}-\bar{T}_{amb})=\theta_{\textit{HDD}} \textit{HDD}.
\label{eq: HDD1}
\end{equation} 

\noindent The difference between the base temperature and the daily average temperature is called Heating Degree Days (HDD). The coefficient $\theta_{\textit{HDD}}$ can be found with linear regression. Similarly a model with Cooling Degree Days (CDD) can be defined to quantify the cooling consumption of a building based on a separate base temperature $T_{amb,b}$. (Depending on the selection of the different base temperatures, the same day can have Heating Degree Days and Cooling Degree Days). While this expression might be a good assumption for most common buildings, it might not be for buildings with large windows, such as the UMAR apartment, where the solar irradiation is a main driver of the dynamics. We therefore add the global solar irradiation $I_{hor}$ with a new regression coefficient $\theta_{sol}$ to the equation,

\begin{equation}
Q_{hea}=\theta_{\textit{HDD}}\textit{HDD}+\theta_{sol}I_{hor}+c,
\label{eq: HDD2}
\end{equation}

\noindent and add a constant $c$ to omit the dependence on the base temperature $T_{amb,b}$.\footnote{If the regression chooses to set $c=-\theta_{HDD} T_{amb,b}$, the function effectively is not dependent on $T_{amb,b}$ any more. We could also write \eqref{eq: HDD2} directly in terms of $\bar{T}_{amb}$, but choose not to, to maintain the relation to the HDD concept.} Finally, we divide the regression coefficients by each other and define the Heating Degree Solar Day (HDSD):

\begin{equation}
Q_{hea}=\theta_{\textit{HDD}}(\textit{HDD}+\frac{\theta_{sol}}{\theta_{\textit{HDD}}}I_{hor})+c=\theta_{\textit{HDD}}\textit{HDSD}+c.
\label{eq: HDSD}
\end{equation}

\noindent The concept of Cooling Degree Solar Days (CDSD) works accordingly, approximating the cooling demand $Q_{coo}$ of a building.

\subsubsection{Experiment results}

\begin{figure*}
	\begin{subfigure}{0.44\textwidth}
		\centering
		\includegraphics[width=1\textwidth]{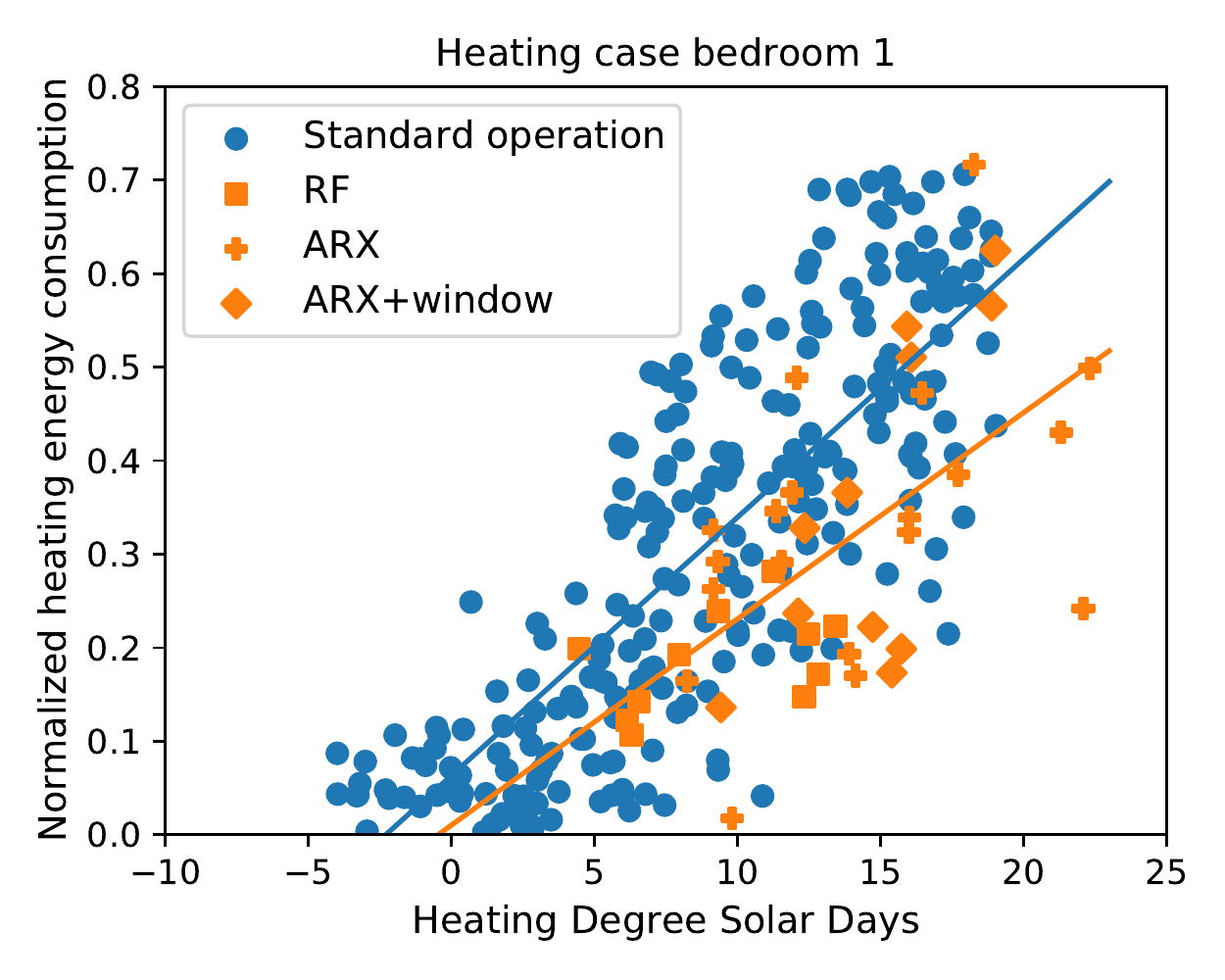}
		\caption{Heating case bedroom 1.}
		\label{fig: Heating br1} 
	\end{subfigure}
	\hfill
	\begin{subfigure}{0.44\textwidth} 
		\centering  
		\includegraphics[width=1\textwidth]{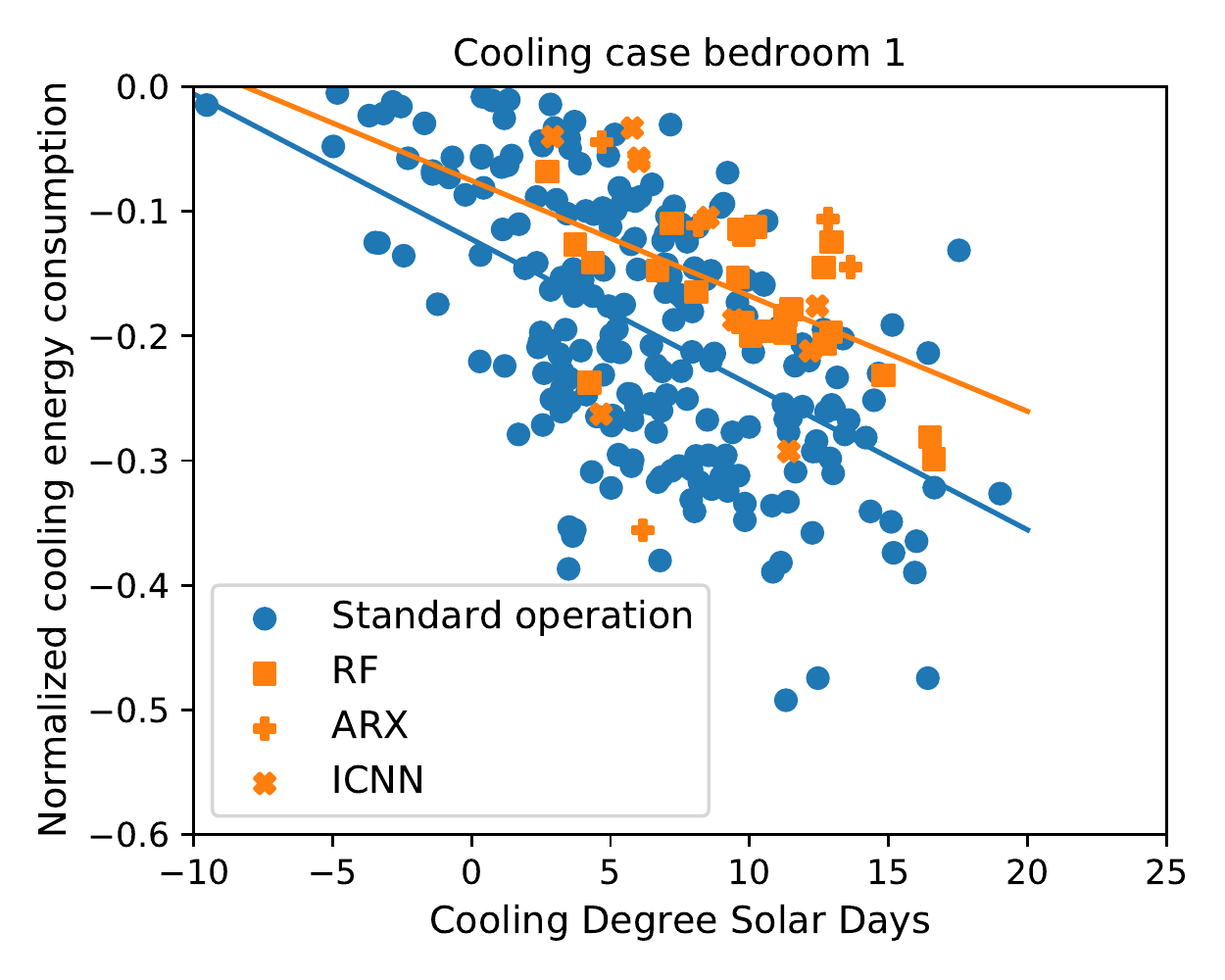}
		\caption{Cooling case bedroom 1.}
		\label{fig: Cooling br1} 
	\end{subfigure}
	\bigskip
	\begin{subfigure}{0.44\textwidth}
		\centering 
		\includegraphics[width=1\textwidth]{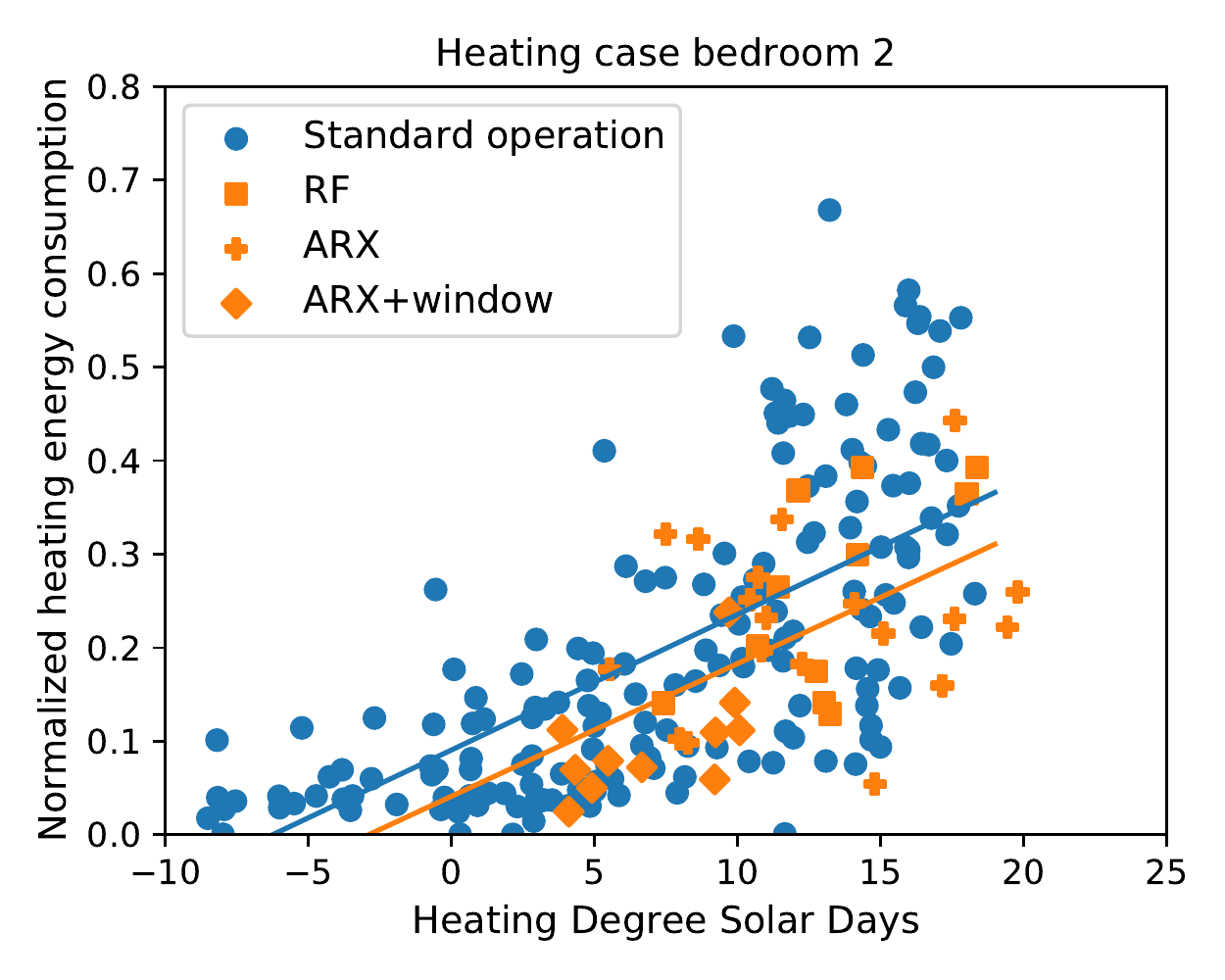}
		\caption{Heating case bedroom 2.}
		\label{fig: Heating br2} 
	\end{subfigure}
	\hfill
	\begin{subfigure}{0.44\textwidth}
		\centering
		\includegraphics[width=1\textwidth]{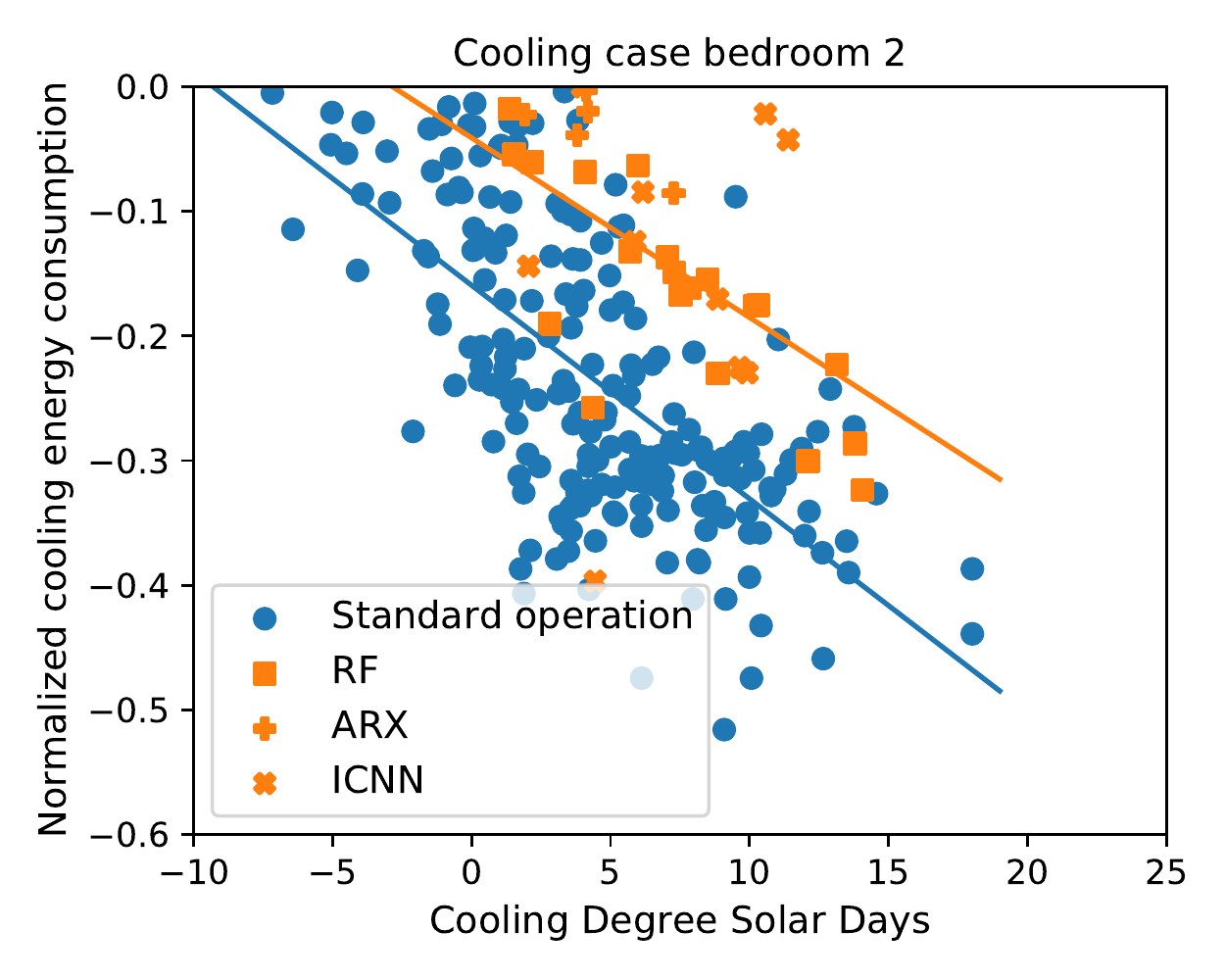}
		\caption{Cooling case bedroom 2.}
		\label{fig: Cooling br2}	
	\end{subfigure}
\caption{Heating energy of MPC methods (orange) and baseline (blue) controller as a function of HDSD and CDSD. Each sample represents one day of experiment, the solid lines show the HDSD and CDSD regressions of these samples. (a): Heating case bedroom 1 with 41 samples for MPC, 258 samples for baseline controller. (b): Cooling case bedroom 1 with 39 samples for MPC, 230 samples for baseline controller. (c): Heating case bedroom 2 with 41 samples for MPC, 184 samples for baseline controller. (d): Cooling case bedroom 2 with 35 samples for MPC, 206 samples for baseline controller.}
\label{fig: energy data}
\end{figure*}

We have evaluated measurement data spanning 2.5 years (2018-06-01 to 2021-02-14) for the analysis of the energy consumption with the MPC controllers compared to the baseline controller (example in Figure \ref{fig: Baseline} in the Appendix).

Figure \ref{fig: Heating br1} shows the heating energy consumption as a function of the HDSD for bedroom 1. Each blue dot represents a day under standard operation with the baseline controller and each orange dot represents a day with a MPC controller. Days where the average room temperature is outside the range of 21 \degree C to 27 \degree C are excluded from the analysis for a fair comparison, as these conditions are often results of the heating/cooling system not working correctly, leading to unrealistically low energy consumption. The lines represent the HDSD regressions. Note, that we have lumped all different MPC models (ARMAX, RF and ICNN)\footnote{The \textit{ARX+window} label denotes a model where eq. \eqref{eq: solar1} is calculated based on the window dimensions and orientations and directly used as a model input instead of eq. \eqref{eq: solar onehot}.} here in orange, as all controllers have shown reasonable control performance in the individual experiments. As the experimental data sets for the individual methods are small, we refrain from distinguishing quantitatively between the modeling methods. However, as we use a different marker for each method, it can be seen that qualitatively, the methods perform similarly.
There is a clear trend visible confirming that the MPC controllers consume less heating energy than the baseline controller. At 5 HDSD the reduction is approximately 41\% while at 15 HDSD it is 29\%. Besides anticipating ``free of cost'' heat gains from solar irradiation and other environmental factors, the MPC controllers of course save a significant amount of energy by exploiting relaxed comfort constraints. However, we note that the energy savings do not stem from a generally substantially lower room temperature set point, or the violation of constraints (as could be seen in Section \ref{subsec: trajectories}). As Figure \ref{fig: Temperature heating br1} in the Appendix shows\footnote{Data on the cooling case in bedroom 1 and both cases in bedroom 2 is available in the linked data repository, see Section \textit{Data Availability}. These data show a similar trend.}, the median daily room temperature difference between MPC and standard operation is 0.18 \degree C. The same trend of MPC consuming significantly less energy compared to the baseline controller is visible in the cooling experiments in bedroom 1, shown in Figure \ref{fig: Cooling br1}. MPC consumes 33\% less cooling energy at 5 CDSD and 28\% less at 15 CDSD.

The trend is slightly less pronounced in the heating experiments in bedroom~2, which are shown in Figure \ref{fig: Heating br2}. Here, the difference between the MPC controllers and the baseline controller is smaller. This behaviour can be explained by the better insulation of bedroom~2. While bedroom~1 has a window surface and a wall surface connected to the ambient, the wall surface of bedroom~2 is adjacent to another unit. This results in less temperature loss during the night, which means that the MPC controller can exploit the relaxed comfort constraint less. The result indicates that the potential for energy savings through MPC depends on the level of insulation of the controlled building, which we are currently investigating in more detail in ongoing simulation-based research. For the cooling experiments in bedroom 2, which are shown in Figure \ref{fig: Cooling br2}, MPC again saves a significant amount of energy compared to the baseline controller. Here, the solar irradiation is the significant disturbance working against the control input, and the window surface area is the same for both bedrooms, which means that relaxed comfort constraints can be exploited in both bedrooms.

When the HDSD and CDSD input data of the entire 2.5 years of measurement data is applied to the linear regression models\footnote{This corresponds to a situation where either the baseline controller or the MPC would have been run for the entire duration of 2.5 years.} in Figure \ref{fig: energy data}, the MPC controllers on average save 33\% and 26\% of heating energy in bedrooms 1 and 2 respectively, and 32\% and 49\% of cooling energy in bedrooms 1 and 2 respectively compared to the baseline controller. Note that, given the deviations from the regressions in Figure \ref{fig: energy data}, this is only a rough estimate. However, it demonstrates that MPC generally saves a significant amount of heating and cooling energy in our experiments, without a significant difference in room temperatures or constraint violations.

%
%
%
%
%

\subsection{Computational performance}

Table \ref{tab: computational resources} summarises the computational requirements for the MPC controllers tested with the different models of this study on a single optimization with fixed initial conditions and disturbance forecasts. The optimization problems were solved on an Intel(R) Core(TM) i7-7500U CPU with 2.7 GHz, and 16 GB of memory. The analysis is done for N=6h, Q=1, R=0, $\lambda$=100.

It can be seen that the ARMAX controller outperforms the other two in terms of solving time and memory usage. While the RF problem is also a QP, the solving time is longer because the parameters for the optimization problem need to be found from the forest on the basis of the non-controllable inputs $X_d$ first. The PICNN has a longer solving time as the optimization problem is more difficult to solve. This is because, even though the problem is convex (without the presence of lower comfort constraints), we do not have access to analytical gradients and the solver has to numerically approximate them. FICNN show similar results. The memory requirements grow with the complexity of the models. While all solving times are acceptable for the considered residential building case, it is to be expected that RF and ARMAX controllers scale better due to their linear dynamics, which could be beneficial for larger scale buildings. We note that all models and solvers are based on open-source libraries, which are not optimized for our specific purpose.

\begin{table}[htb]
\centering
\caption{Computational resources for MPC controllers with varying models.}
\label{tab: computational resources}
\scriptsize
\begin{tabular}{|l|r|r|r|c|l|l|l|l|l|}
\hline
 Model		& Solving time & Memory usage & Software (Python 3) \\ \hline
 ARMAX 		& 0.2 s & 36 MB & Scikit-learn, CVXOPT\\
 RF		& 1.5 s & 68 MB  & Scikit-learn, CVXOPT\\
 PICNN 		& 3.0 s & 158 MB & Keras, SciPy\\ \hline
\end{tabular}
\normalsize
\end{table}

\section{Numerical case study}
\label{sec: Numerical case study}


\begin{figure}
\centering
\includegraphics[width=0.44\textwidth]{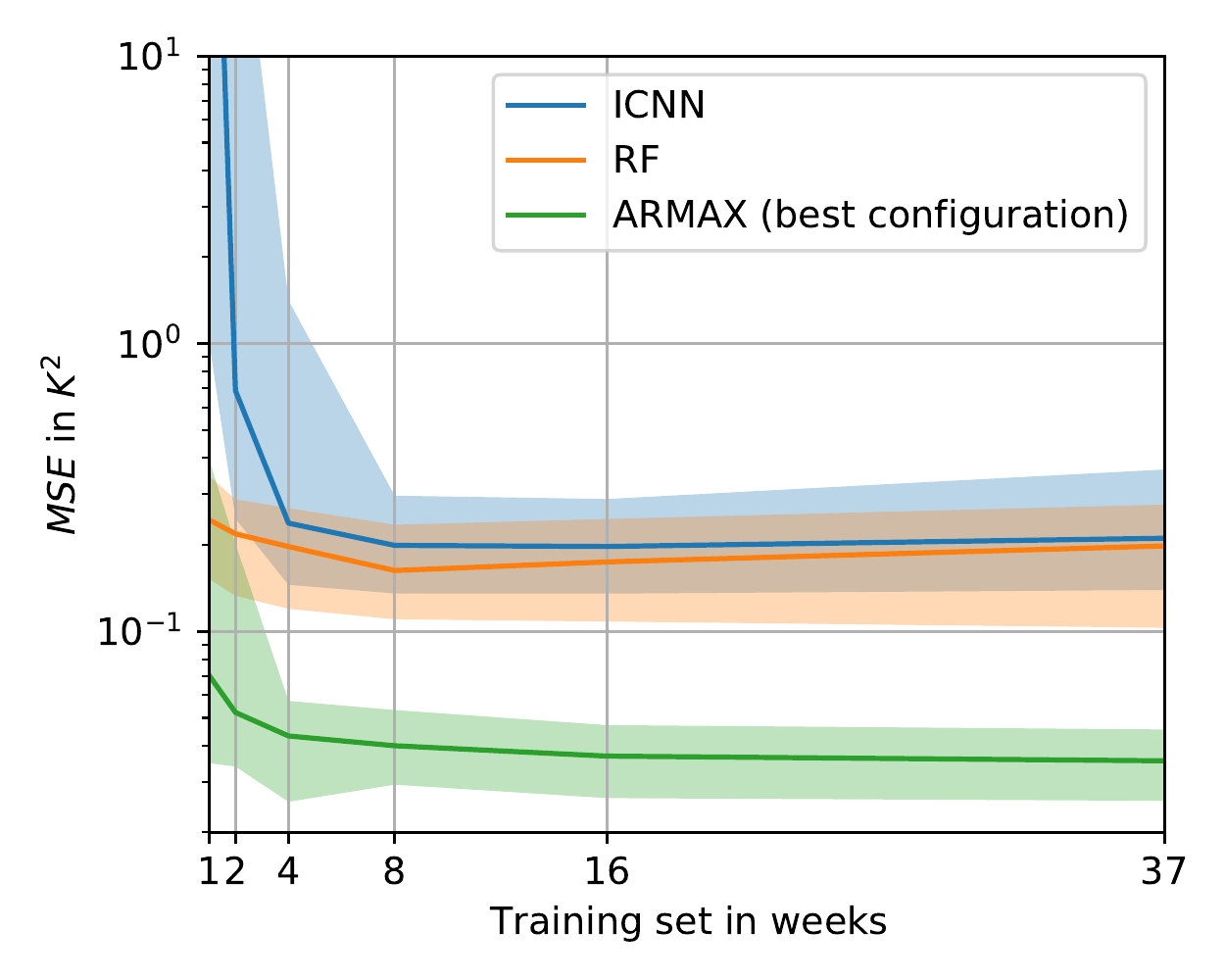}
\caption{Mean Squared Error for 1-hour open loop prediction with ICNN (blue), RF (orange) and ARMAX (green) models. The solid lines depict the median achieved MSE, and the shaded areas depict the 16\% and 84\% percentiles.}
\label{fig: SampleEff conclusion}
\end{figure}

\begin{figure}
\centering
\includegraphics[width=0.44\textwidth]{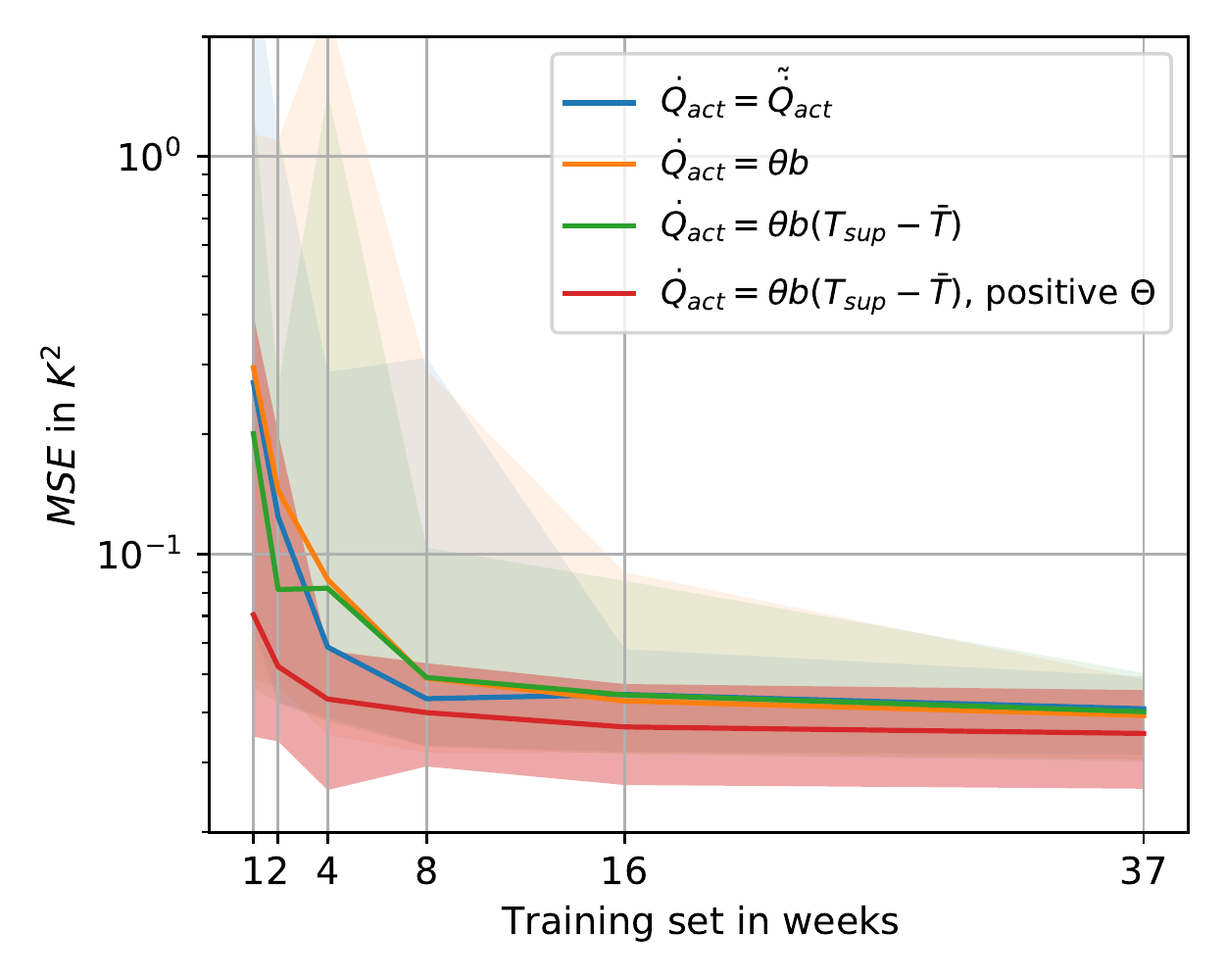}
\caption{Mean Squared Error for 1-hour open loop prediction with ARMAX model and varying inputs. The solid lines depict the median achieved MSE, and the corresponding coloured shadows depict the 16\% and 84\% percentiles.}
\label{fig: SampleEff ARX}
\end{figure}

Our experiments have shown that MPC with all the considered models is generally suitable for building control in practice. However, to minimize costs of commissioning, the amount of historical measurement data required to train the model is a considerable factor for choosing a model for MPC. To address this point, we compare here the sample efficiency of the ARMAX, RF and ICNN models.

For this, the training data from 2018-05-23 to 2019-05-28 for bedroom 1 is divided into weekly folds and a k-fold cross validation is performed, where a randomly selected subset of the data is used for training and the rest is used for model validation in terms of the mean squared error (MSE) for a one-hour open loop prediction. The numerical experiment is repeated 100 times to account for different training and validation data. Figure \ref{fig: SampleEff conclusion} shows the result for a PICNN, RF (both tuned for this particular bedroom), and an ARMAX model with three lag terms\footnote{Our validation experiments suggest that including more than three lag terms does not improve the accuracy of the predictions of the ARMAX model in this case.} using the positivity constraint on the elements of $\Theta$ in \eqref{eq: ARX2}. We do not show results of FICNN as they are very similar to those of PICNN. All models use $\tilde{\dot{Q}}_{act,i}$ (Equation \eqref{eq: Qact true}) as the control input.

It can be seen that the ICNN performs poorly for a single week and two weeks of training data, and significantly improves after four weeks. This is most likely due to the large number of model parameters, which can lead to overfitting. In contrast, the RF already performs relatively well for a single week of training data but does not improve much when more training data is available. The good performance for a small training data set is due to the automatic scaling of RF through the minimum number of samples per leaf. The median of the ARMAX model outperforms both the ICNN and RF models for all sizes of training data. This could be expected for small training data sets as the physical priors lead to strong regularization. However, it can also be seen that the ICNN and RF models converge to the same median MSE and similar variance, while the ARMAX model's median converges to a much lower MSE and less variance. This indicates that using physical model inputs gives the ARMAX model domain knowledge that the ML models cannot find by themselves (for example the solar irradiation through windows as a function of time and global solar irradiation), even if abundant training data is available. 


As the number of parameters of the ARMAX model is similar to the RF model for low-sized training sets, the higher variance of the prediction error of the ARMAX model for one and two weeks of training data, is most likely an effect of the one-hot encoding of $I_{vert(t)}$ (the vertical solar irradiation): for small training sets it can happen that the set does not include samples from the summer, which means that $I_{vert,1}$ and $I_{vert,\tau}$ are always zero in the training set because the sun rises late and sets early\footnote{In our implementation $I_{vert,0}$ covers the night-time, therefore $I_{vert,\tau}$ (late evening) and $I_{vert,1}$ (early morning) may or may not be non-zero depending on the season.}. Accurate coefficients can therefore not be found during training. If samples from the summer are included in the validation set, this leads to large prediction errors. For real applications, the issue is likely less pronounced, if the models are updated regularly. The issue could also be addressed through further regularization, e.g.~with the Lasso method \cite{Tibshirani1996}. 

For practical implementations of data-driven MPC, it is important to know which quantities need to be measured in a building. We therefore compare ARMAX models with various choices for the control input, as described in Section \ref{subsec: actuator gains}, in Figure \ref{fig: SampleEff ARX}. The first model (blue) uses the considered \textit{true} energy input $\tilde{\dot{Q}}_{act}$ (Equation \eqref{eq: Qact true}), the second one (orange) only uses the valve opening $b$ (Equation \eqref{eq: act2}), and the third one (green) uses the approximation via valve opening, supply temperature $T_{sup}$ and a previously measured room temperature $\bar{T}$ (Equation \eqref{eq: act3}). The last model (red) additionally constrains all regression coefficients $\Theta$ in \eqref{eq: ARX2} to be non-negative. The first three models perform similarly well, which indicates that using just the valve opening to model the control input could be sufficient in practical cases - an observation that is also supported by our experiments. Measuring mass flows and supply temperatures gives no visible performance advantage, which simplifies practical implementation. The red result in Figure \ref{fig: SampleEff ARX} demonstrates that the positivity constraint on $\Theta$ significantly benefits the sample efficiency. This is especially important for practical implementation, as it reduces controller commissioning time. The model variance and the median MSE are also reduced. 

Although the constraints on $\Theta$ increase the prediction accuracy, they do not guarantee BIBO stability. For the {ARMAX} model, stability could be guaranteed by further constraining the coefficients of the auto-regressive inputs, which is part of ongoing research. Ensuring stability for ML models such as the RF and ICNN models is much more complex and not appropriately addressed in the literature so far. Here, methods such as presented by \cite{Drgona2020a} for bounding the dominant eigenvalues of a Recurrent Neural Network based building thermal model should be further explored.
As also supported by our experiment results in Section \ref{sec: Experiment case study}, in practice, as buildings have very well-damped stable physics, models identified from their data are also usually stable if the data quality is reasonable. This can be confirmed, for example, by observing the coefficients of the autoregressive inputs of the ARMAX models trained on different size training sets, as done in this work, and different types of buildings (see also \citep{Lefebure2021}, where the ARMAX method is applied to a floor-heating-based medium weight construction, in contrast to the radiator-based light-weight construction modeled in this paper).

\section{Conclusion}
\label{sec: Conclusion}

In this paper, we have compared physics-informed ARMAX models to Machine Learning based models in the domain of MPC for building climate control in experiments and in numerical case studies.

The study has shown that MPC with all three models, RF, ICNN and ARMAX, generally delivers good control performance. Moreover, in all heating and cooling cases, MPC achieves significant energy savings compared to the baseline controllers. Our results also suggest that the physics-informed ARMAX models outperform the RF and ICNN models in terms of online computational requirements and offline training sample efficiency, which means that good models can be extracted from less data. The latter finding suggests that the physics-based inputs and constraints give the model an information prior, which cannot be found by the ML methods themselves, even if abundant training data is available. The increased expressiveness of ML-based models therefore does not seem to add any benefits in this case. Together with the lower computational requirements in terms of optimization time and memory usage, the results render the ARMAX approach superior for practical implementation compared to the ML methods.

The next logical step is to apply the physics-informed inputs to the ML methods and to find ways to enforce physics-based constraints, such as non-negativity, with ANN and RF. Physics-informed ML methods could potentially be interesting as soon as non-linearities, for example heat pumps, are added directly to the control problem, instead of being treated on a lower control level. Moreover, experiments with the physics-informed ARMAX models should be conducted on large-scale buildings. This is currently under investigation with an industry partner.

\section*{Data Availability}
\label{sec: Data Availability}

The data presented in this work, and data from additional experiments is available under the DOI 10.3929/ethz-b-000496285.

\section*{Acknowledgement}

We would like to thank Kristina Orehounig for her valuable help and support. We are also grateful to Varsha Behrunani, Annika Eichler, Benjamin Flamm, Marta Fochesato, Andrea Iannelli, Mohammad Khosravi, Francesco Micheli, Anil Parsi and Joseph Warrington for fruitful discussions. We particularly would like to thank Reto Fricker, Ralf Knechtle and Sascha Stoller for sharing their expertise and their help with the implementation, and Antoon Decoussemaeker for accompanying research.

This research project is financially supported by the Swiss Innovation Agency Innosuisse and is part of the Swiss Competence Center for Energy Research SCCER FEEB{\&}D. It is also financially supported by the SNSF under the NCCR Automation.

\section*{Declaration of competing interest}

The authors declare that they have no known competing financial interests or personal relationships that could have appeared to influence the work reported in this paper.

\printcredits

\appendix
\section{Appendix}

\begin{figure}[ht]
\centering
\includegraphics[width=0.44\textwidth]{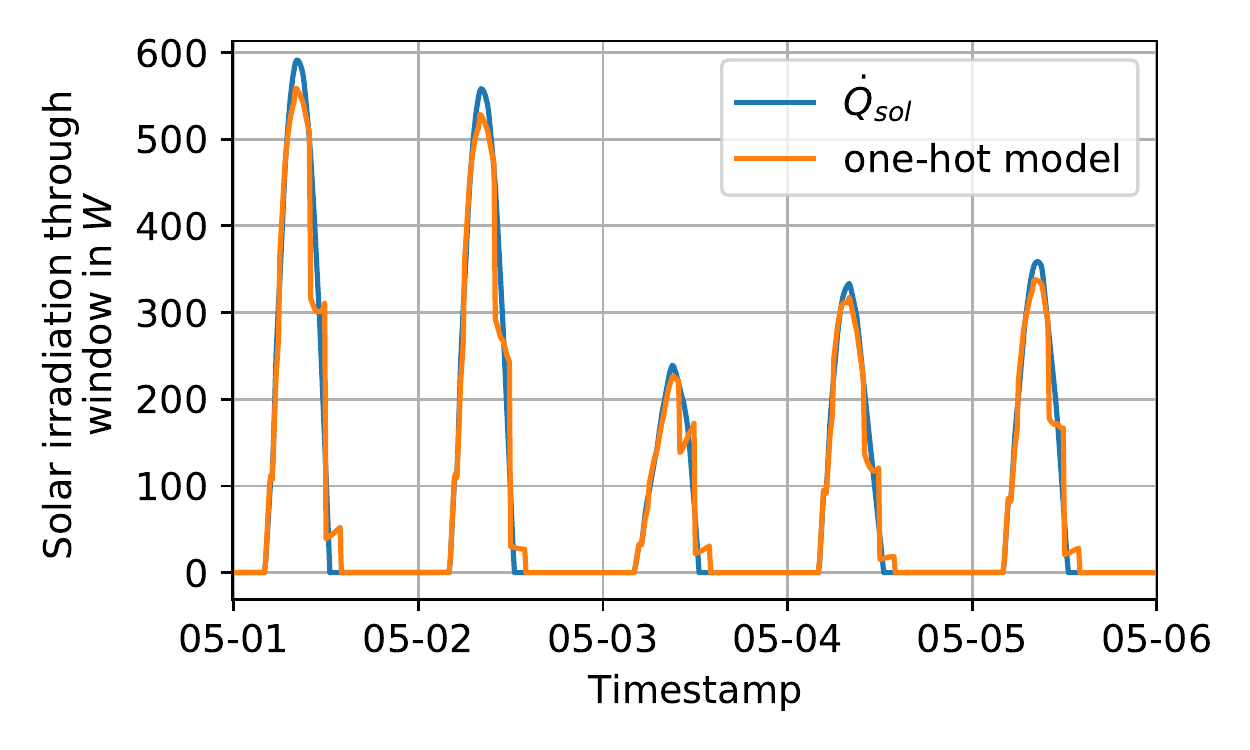}
\caption{Example from the validation set of the one-hot solar model. The blue line denotes the \textit{true} solar gains through a window. The orange line denotes the predicted gains by the model.}
\label{fig: Sol valid}
\end{figure}

\begin{figure*}
\centering
\includegraphics[width=1\textwidth]{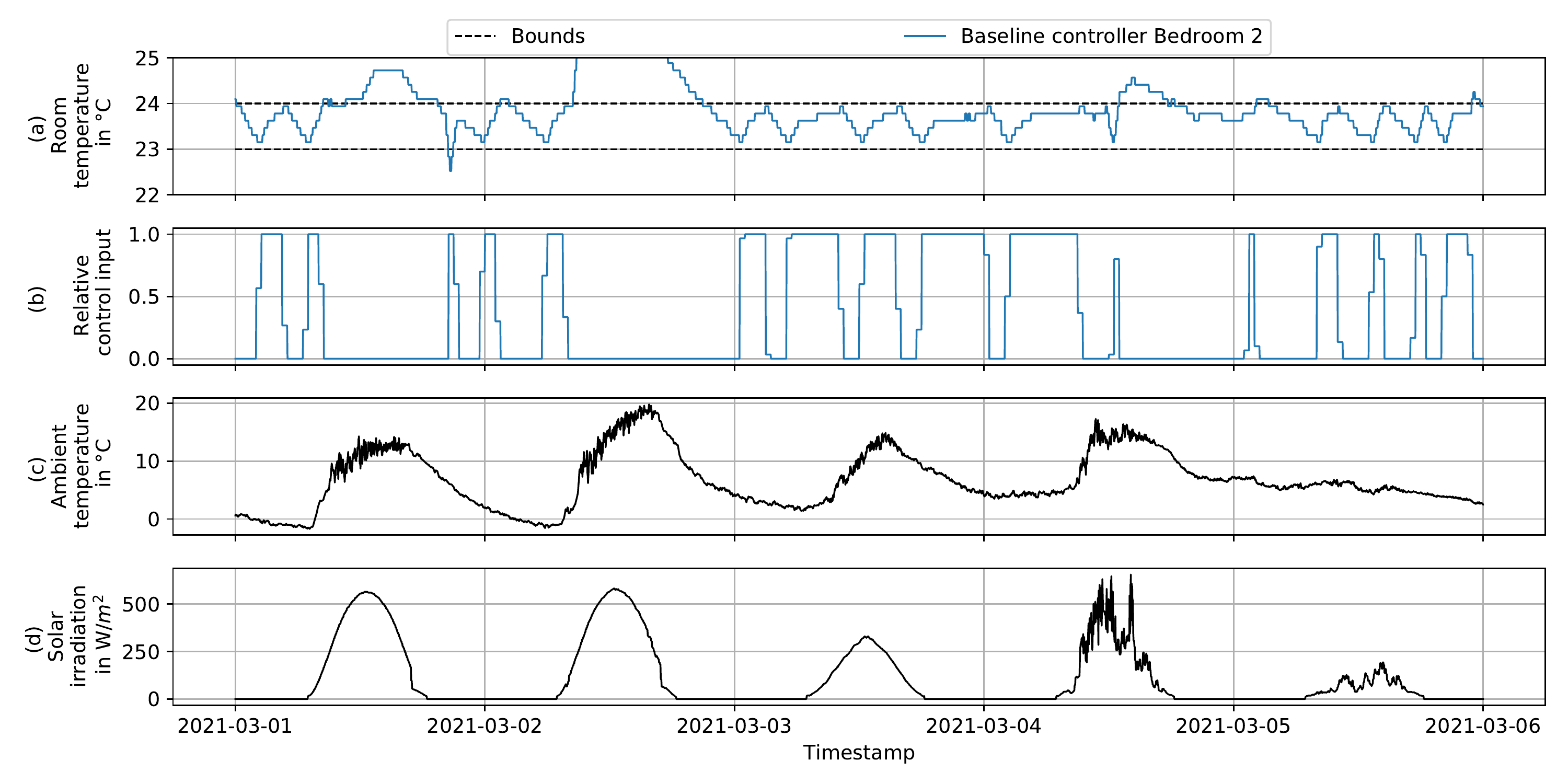}
\caption{Baseline controller example for the heating case in bedroom 2. (a): Temperature in the bedroom. The limits of the hysteresis controller are shown in dashed black. (b): Relative control input, i.e. the fraction of time where the maximum control input is applied during one control step. (c): Measured ambient temperature at the experiment site. (d): Global solar irradiation at the experiment site.}
\label{fig: Baseline}
\end{figure*}

\begin{figure}[ht]
\centering
\includegraphics[width=0.44\textwidth]{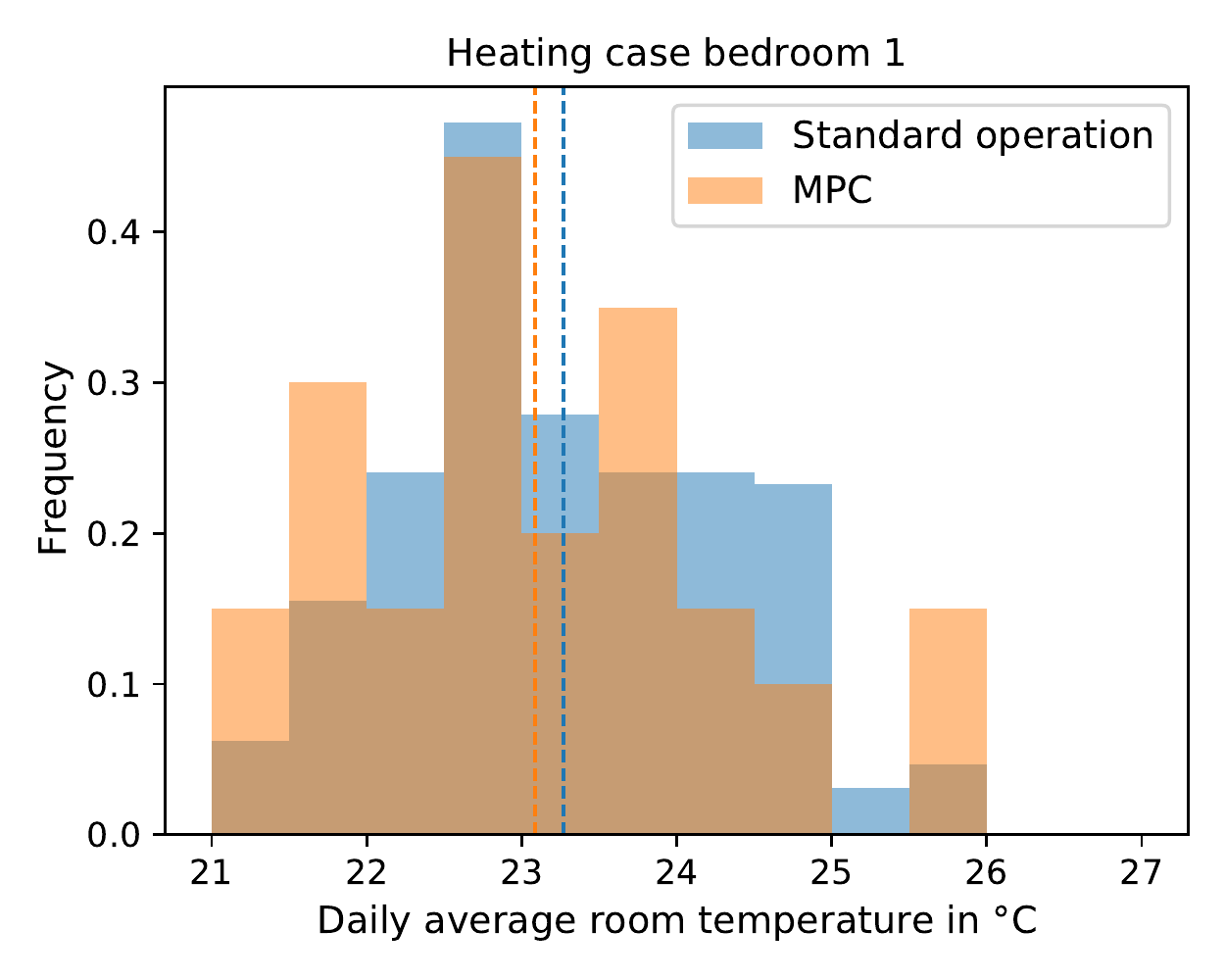}
\caption{Temperature analysis MPC vs. baseline controller for heating experiments in bedroom 1.}
\label{fig: Temperature heating br1}
\end{figure}

See Figures \ref{fig: Sol valid} - \ref{fig: Temperature heating br1}.

%
%

%
%
%

\end{document}